\documentclass[10pt,twocolumn,letterpaper]{article}

\usepackage{cvpr}              

\definecolor{cvprblue}{rgb}{0.21,0.49,0.74}
\usepackage[pagebackref,breaklinks,colorlinks,allcolors=cvprblue]{hyperref}
\usepackage{booktabs}
\usepackage{multirow}
\usepackage{wrapfig}
\usepackage{caption}
\usepackage{algorithm}
\usepackage{algorithmicx}
\usepackage{algpseudocode}
\floatname{algorithm}{Algorithm}

\usepackage{array}

\def\paperID{274} 
\def\confName{CVPR}
\def\confYear{2025}

\title{Towards General Visual-Linguistic Face Forgery Detection}

\author{   
\normalsize Ke Sun$^{1}$, Shen Chen$^{2}$, Taiping Yao$^{2}$, Ziyin Zhou$^{1}$, Jiayi Ji$^{1}$, Xiaoshuai Sun$^{1}$\thanks{Corresponding author}, Chia-Wen Lin$^{3}$, Rongrong Ji$^{1}$ \\
\normalsize $^1$ Key Laboratory of Multimedia Trusted Perception and Efficient Computing, \\
\normalsize Ministry of Education of China, Xiamen University \\
\normalsize $^2$ Youtu Lab, Tecent, China\\
\normalsize $^3$ National Tsing Hua University, Taiwan
}

\begin{document}
\maketitle

\begin{abstract}

Face manipulation techniques have achieved significant advances, presenting serious challenges to security and social trust. Recent works demonstrate that leveraging multimodal models can enhance the generalization and interpretability of face forgery detection. However, existing annotation approaches, whether through human labeling or direct Multimodal Large Language Model (MLLM) generation, often suffer from hallucination issues, leading to inaccurate text descriptions, especially for high-quality forgeries. To address this, we propose Face Forgery Text Generator (FFTG), a novel annotation pipeline that generates accurate text descriptions by leveraging forgery masks for initial region and type identification, followed by a comprehensive prompting strategy to guide MLLMs in reducing hallucination. We validate our approach through fine-tuning both CLIP with a three-branch training framework combining unimodal and multimodal objectives, and MLLMs with our structured annotations. Experimental results demonstrate that our method not only achieves more accurate annotations with higher region identification accuracy, but also leads to improvements in model performance across various forgery detection benchmarks.
Our Codes are available in \href{https://github.com/skJack/VLFFD.git}{https://github.com/skJack/VLFFD.git.}
\end{abstract}

\section{Introduction}
\label{sec:intro}
Face manipulation techniques have achieved remarkable progress in recent years, enabling high-quality modifications of facial attributes~\cite{gonzalez2018facial}, expressions~\cite{liu2019stgan}, and identities~\cite{korshunov2018deepfakes}. While these advances bring creative possibilities, they also raise serious concerns about potential misuse and social trust~\cite{tolosana2020deepfakes}. To address these challenges, developing robust face forgery detection methods has become crucial, especially for handling unseen forgeries that exhibit significant domain gaps from training data~\cite{sun2021dual,sun2021domain,lin2024preserving}.

\begin{figure}[t]
    \begin{center}
       \includegraphics[width=1\linewidth]{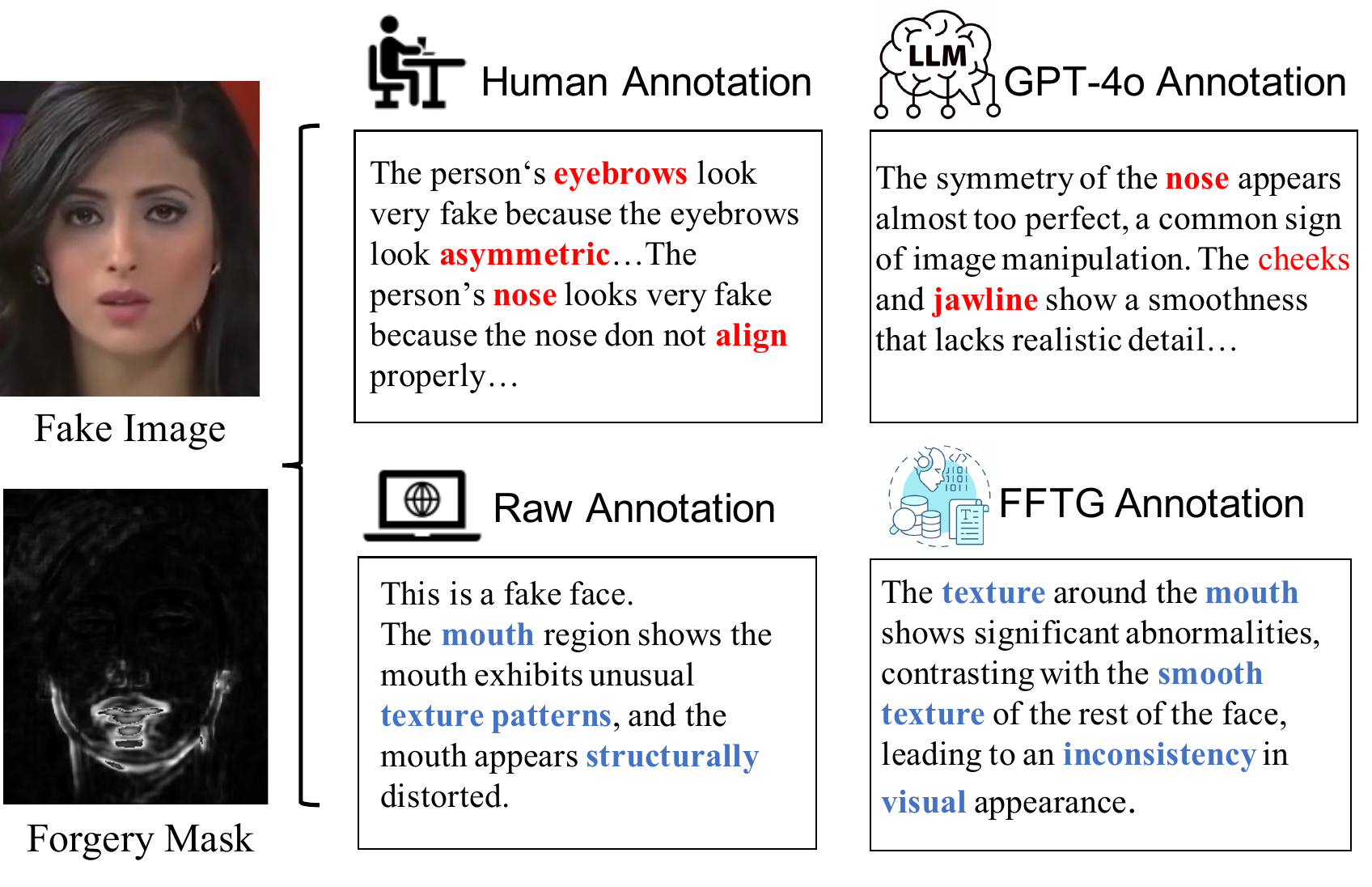}
    \end{center}
       \caption{ Differences between annotations generated by human annotation~\cite{zhang2024common}, GPT-4o methods and ours for a fake image. The fake image is manipulated only on the mouth region, and the forgery mask is generated by comparing the difference between real and fake images.
       (Best viewed in color.)
       }
    \label{fig:intro}
\end{figure}

Most existing face forgery detection methods rely on unimodal architectures, which often lack interpretability and generalization. Recent advancements in visual-language multimodal learning, such as CLIP~\cite{radford2021learning} and multimodal large language models (LLMs), have demonstrated powerful representation learning capabilities for both visual and language tasks. These models create a bridge between vision and language, improving human understanding of visual tasks and enhancing model performance through language-guided learning. For face forgery detection, incorporating language modality could provide interpretable explanations and tap into the rich semantic knowledge embedded in multimodal models~\cite{zhang2024common,huang2024ffaa,liu2024evolving,chen2024textit}.

To leverage these powerful multimodal models for forgery detection, high-quality text annotations are essential.
However, obtaining accurate text annotations for face forgery data remains challenging. Current approaches for obtaining text annotations primarily fall into two categories: \textit{Human Annotation}~\cite{zhang2024common}, where annotators manually identify forgery regions and provide explanations, and \textit{MLLM Annotation}~\cite{huang2024ffaa}, where prompts are crafted to enable multimodal large language models (e.g., GPT-4o) to generate annotations. 
However, we have observed that both approaches suffer from \textit{hallucination issues}, especially for high-quality forged faces.
For instance. as shown in Figure \ref{fig:intro}, 
we visualize the annotations of NeuralTexture forgeries in the FFpp~\cite{rossler2019faceforensics++} dataset produced by DD-VQA~\cite{zhang2024common} and GPT-4o. The forgery is limited to the mouth region, while other regions are authentic. Both human and MLLM annotations incorrectly mark the nose area, which remains unaltered in the forged image. Such annotation errors impact the performance and interpretability of downstream tasks.

To address these challenges, we propose a data annotation pipeline called \textbf{Face Forgery Text Generator (FFTG)}, which mitigates hallucination by incorporating accurate forgery region localization and type identification as concrete guidance for text generation. Specifically, FFTG first generates forgery maps by comparing real and forged images, assesses the forgery degree of each facial component, and uses handcrafted features to estimate forgery types, combining these elements into a raw annotation. We then design a comprehensive prompting strategy to guide multimodal large language models (e.g., GPT-4o mini) in generating accurate annotations. Our strategy consists of 1) paired real-fake images as visual prompts enabling the model to identify differences through comparison, 2) guide prompts containing the raw annotation and its derivation process to reduce hallucination, 3) task description prompts that guide the model to perform step-by-step analysis through chain-of-thought reasoning, and 4) pre-defined prompts that standardize output format and provide additional guidelines. As shown in Figure \ref{fig:intro}, this carefully designed pipeline produces more accurate and diverse annotations compared to existing methods.

We validate the effectiveness of FFTG-generated annotations by fine-tuning both CLIP and multimodal LLMs (e.g., LLaVA). For the CLIP model, we adopt a multimodal joint training approach, aligning and integrating the text and visual modalities to assist classification, allowing the text to better guide the visual encoder. Experimental results demonstrate that FFTG annotations enable better generalization performance compared to traditional methods when fine-tuning CLIP. For multimodal LLMs, our annotations not only provide better interpretability but also achieve higher accuracy compared to human annotations and direct GPT labeling. This indicates that the detailed and structured prompts in FFTG reduce annotation errors, resulting in improved model performance across various metrics.

Our main contributions can be summarized as follows:
\begin{itemize}
    \item We identify a fundamental challenge in visual-linguistic forgery detection: obtaining accurate text annotations that align with forgery masks.
    
    \item We propose FFTG, a novel annotation pipeline that leverages forgery masks to generate accurate and diverse text annotations for deepfake images.
    
    \item We demonstrate the effectiveness of our annotations through extensive experiments with CLIP and MLLM, showing improved generalization and interpretability.
\end{itemize}
\section{Related Work}
\label{sec:related}
\subsection{General Face Forgery Detection}

General face forgery detection focuses on improving model generalization to unseen domains, which remains a critical challenge in this field. Existing approaches mainly fall into two categories: forgery simulation and framework engineering. The former simulates various forgery traces through data augmentation, including blending artifacts~\cite{guillaro2023trufor,li2020face,shiohara2022detecting}, facial inconsistencies~\cite{zhao2021learning,chen2022self,sun2024diffusionfake,yan2024transcending,tan2024rethinking}, and subtle distortions~\cite{larue2023seeable,nguyen2024laa}. The latter enhances network architectures through attention mechanisms~\cite{zhao2021multi,xu2023tall,shao2023detecting,sun2022information}, frequency-spatial modeling~\cite{qian2020thinking,luo2021generalizing,liu2021spatial}, or implicit identity modeling~\cite{huang2023implicit,dong2023implicit,choi2024exploiting}. Recent works also explore local-global relationships~\cite{wang2023dynamic,guillaro2023trufor} and feature disentanglement~\cite{dong2022protecting,yan2023ucf,hong2024contrastive,sun2024continual} for better generalization. However, these methods ignore the fine-grained semantic information, which can help the model obtain more generalization features.

\subsection{Visual-Language Learning on FFD}

The visual-language pretraining paradigm, such as CLIP~\cite{radford2021learning} through multimodal contrastive learning, has recently been extended to face forgery detection. Early attempts like DD-VQA~\cite{zhang2024common} utilized crowdsourcing platforms to collect human annotations for deepfake data and fine-tuned multimodal models like BLIP~\cite{li2022blip}. With the advancement of multimodal large language models (MLLMs), researchers began exploring their capabilities in forgery detection. \citet{jia2024can} first investigated GPT's ability in detecting manipulated faces, while FFAA~\cite{huang2024ffaa} leveraged GPT-4o for annotation generation and model fine-tuning. X2DFD~\cite{chen2024textit} further proposed a self-enhancement approach for improving MLLM performance in forgery detection. However, the effectiveness of these methods heavily relies on annotation quality. Our work addresses this fundamental challenge by providing a more accurate annotation pipeline that leverages concrete visual evidence to guide text generation.

\section{Face Forgery Text Generator}
\label{sec:fftg}

\begin{figure*}[!t]
    \begin{center}
    
       \includegraphics[width=1.0\linewidth]{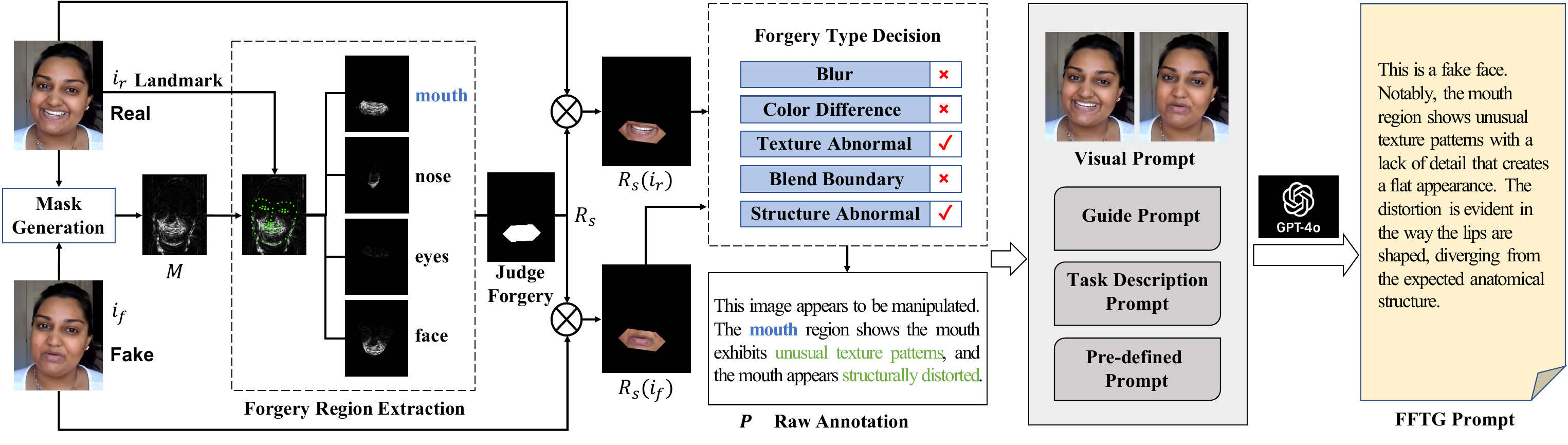}
    \end{center}
       \caption{ Overall framework of the Face Forgery Text Generator (FFTG). The paired forgery and real image are first fed into the Mask Generation module to generate forgery mask $M$. Then the Forgery Region Extraction module extracts the selected region $R_s$. Subsequently, the Forgery Type Decision module decides the forgery type and generates raw annotation. Then the final annotation is generated by GPT with several prompts.
       }
    \label{fig:fftg}
\end{figure*}
\begin{figure}[!t]
    \begin{center}
    
       \includegraphics[width=1\linewidth]{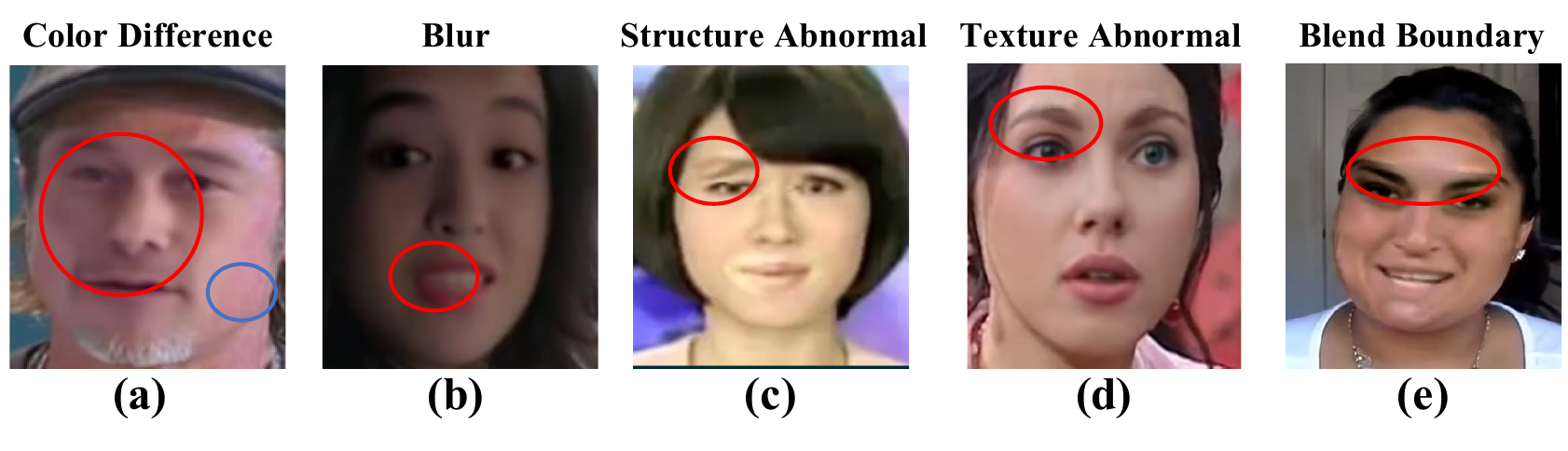}
    \end{center}
       \caption{Five typical types of forgery faces. (a) Color Difference. (b) Blur. (c) Structure Abnormal.
       (d) Texture Abnormal. (e) Blend Boundary.
       The red circle highlights the region of each forgery type. (Best viewed in color.)
       }
    \label{fig:type}
\end{figure}

In this section, we introduce our proposed FFTG pipeline, which comprises the \textbf{Raw Annotation Generation (RAG)} and \textbf{Annotation Refinement}. The goal of RAG is to provide an initial annotation using handcrafted criteria and accurate forged images. Although the generated annotations are limited in diversity and have a relatively fixed structure, they are highly accurate and reasonable, which helps to reduce the hallucinations that may occur when using large language models for annotation. Annotation Refinement with MLLM leverages advanced multimodal large language models (e.g., GPT-4o-mini) to further refine the annotations. To increase diversity and improve accuracy, we employ four types of prompts to guide the large model in this process. The overall framework is shown in Figure \ref{fig:fftg}.

\subsection{Raw Annotation Generation}
Given a real image $i_r \in \mathbb{R}^{ 3\times H \times W}$ and its corresponding forgery image $i_f \in \mathbb{R}^{ 3\times H \times W}$,  RAG encompasses the following steps:

    \noindent\textbf{Mask Generation}. To locate the forgery region, similar to~\cite{chen2021local}, we first construct manipulated mask $M$ by computing the absolute pixel-wise difference in the RGB channels, and then normalizing it to the range of $[0,1]$:
    \begin{equation}
        M = |i_r-i_f| / 255.
        \label{equation:1}
    \end{equation}
    
    \noindent\textbf{Forgery Region Extraction}. 
    This step aims to select a forgery region containing $i_f$. Facial images are divided into four areas: mouth, nose, eyes, and face, based on landmarks. We compute the average value of $M$ in each area and set a threshold $\theta$ to form the forgery region list $L_f$. This is defined as:
    \begin{equation}
        \frac{1}{|R_t|}\sum_{j\in R_t}M(j) > \theta, R_t \rightarrow L_f,
        \label{equation:2}
    \end{equation}
    where $R_t$ represents one of the four predefined areas, and $|R_t|$ is the sum of pixels in area $t$. If the value exceeds $\theta$, the corresponding area is included in $L_f$. After processing all four areas, we randomly select one region $R_s$ from $L_f$ for the next step. $R_s(i_r)$ and $R_s(i_f)$ represent the forgery regions for real and fake pixels, respectively.
    
    \noindent\textbf{Forgery Type Decision}. The goal of this step is to determine the type of forgery via a specially designed criterion.
    According to the previous work and our observation, we categorize the existing forgery types as color difference, blur, structure abnormal, texture abnormal, and blend boundary as shown in Fig~\ref{fig:type}. We detail each forgery type and corresponding evaluation standard as follows:
    1) \textbf{Color Difference }: Occurs in face swaps with notable color variance. We assess this using the distance of average channel-wise mean and variance in $Lab$ color space between real and fake regions.
    2) \textbf{Blur}: We use the Laplacian operator to quantify local blurring in forgery faces, determining blurriness by the variance after applying the operator to real and fake images in the selected region.
    3) \textbf{Structure Abnormal}: Observed deformations in fake face organs are assessed using the SSIM index difference between real and fake images in the selected region $R_s$.
    4) \textbf{Texture Abnormal}: We measure texture clarity using the contrast of the Gray-Level Co-occurrence Matrix (GLCM), defining an area as texture abnormal when the real region's $C_d$ exceeds that of the fake beyond a threshold.
    5) \textbf{Blend Boundary}: Existing face manipulation methods conduct blending operation to transfer an altered face into an existing background, which leaves intrinsic cues across the blending boundaries~\cite{li2020face}, such as the red circle of Figure.~\ref{fig:type}(e). We assess the presence of blending artifacts by analyzing three characteristics in the selected region's boundary: gradient variations, edge transitions, and frequency domain changes. If at least two of these metrics exceed their respective thresholds, we classify the region as having significant blending boundaries.

Supplementary materials provide detailed pseudocodes for each criterion. The identified regions and types are then transformed into natural language expressions using GPT-4o generated descriptive phrases. For instance, ``Texture Abnormal" becomes ``lacks natural texture" and ``Color Difference" translates to ``has inconsistent colors". A complete list of these mappings is provided in the supplementary materials. This translation ensures our raw annotations are both technically accurate and linguistically natural, facilitating subsequent refinement by MLLMs.

\subsection{Annotation Refinement with MLLM}


While our mask-guided analysis provides accurate region localization, the handcrafted features may not fully capture all forgery types, and the generated descriptions lack linguistic diversity. To address these limitations, we leverage GPT-4o mini's strong visual understanding capabilities for refined annotation generation. To ensure both accuracy and diversity while avoiding hallucination, we design a comprehensive prompting strategy with four key components:

\noindent\textbf{Visual Prompt}: Instead of presenting single images, we concatenate the real and forged face images as paired inputs to the MLLM. This comparative approach serves two purposes: 1) enables the model to identify forgery artifacts through direct comparison, reducing hallucination by providing explicit visual references, and 2) helps generate more focused annotations for real images by maintaining the forgery detection perspective, avoiding irrelevant descriptions that might emerge from isolated real image.


\noindent\textbf{Guide Prompt}: We incorporate the previously generated raw annotations into this component, along with detailed explanations of how each forgery type was determined. For example, we explain how texture abnormalities were identified using GLCM analysis and how structural deformations were determined through SSIM comparisons.

\noindent\textbf{Task Description Prompt}: Clear instructions establish an expert forgery detection context, defining specific requirements for analyzing visual evidence and generating comprehensive descriptions of manipulation artifacts.

\noindent\textbf{Pre-defined Prompt}: Structured output requirements specify JSON format, mandatory phrases (``This is a real/fake face"), and caption counts for consistent annotation generation, ensuring standardized outputs for downstream tasks.

This strategy enables the model to generate accurate and diverse annotations while maintaining consistency with technical analysis. Due to space limitations, we provide the complete prompt templates in the supplementary material.

\begin{figure}[t]
    \begin{center}
       \includegraphics[width=1\linewidth]{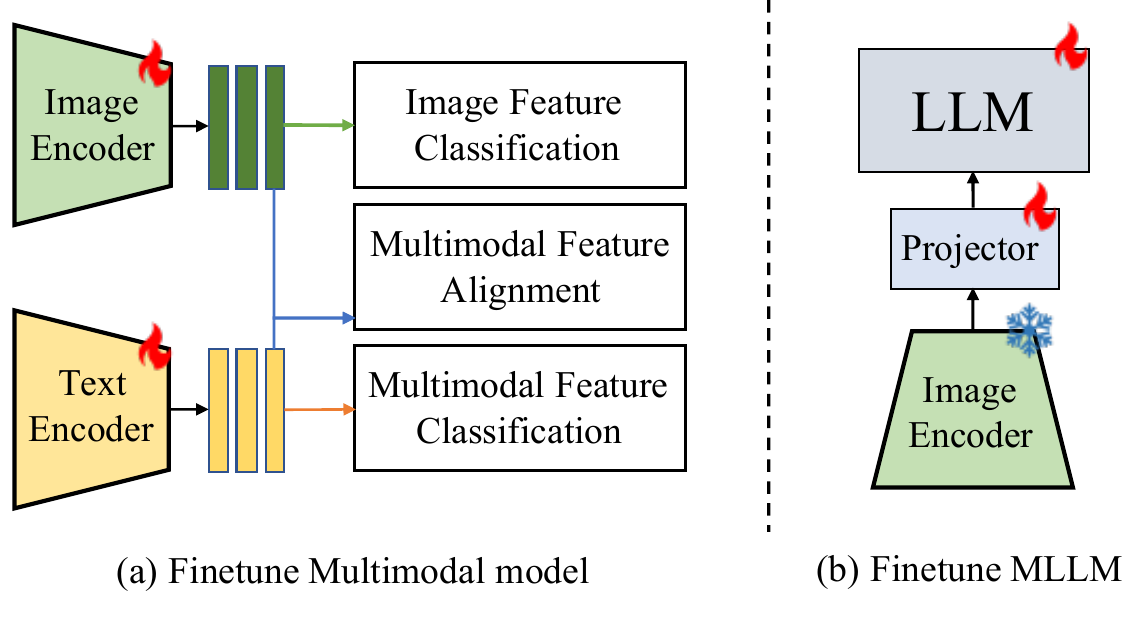}
    \end{center}
       \caption{
       Overview of our fine-tuning strategies. (a) For multimodal models like CLIP, we employ three training objectives: direct image classification, feature alignment between modalities, and multimodal fusion classification. (b) For MLLM, we utilize our pre-trained image encoder and fine-tune the projector and LLM components.
       }
    \label{fig:finetune}
\end{figure}

\section{Model Fine-tuning}
\label{sec:finetune}

To better validate the effectiveness of FFTG for face forgery detection, we provide two baseline approaches for utilizing our annotations, as illustrated in Figure~\ref{fig:finetune}. The first baseline focuses on fine-tuning multimodal models like CLIP through a three-branch learning framework that combines both unimodal and multimodal objectives. The second baseline explores enhancing multimodal large language model (MLLM), aiming to improve both their forgery detection accuracy and reasoning capabilities.

\subsection{Finetune Multimodal Models}

As shown in Figure~\ref{fig:finetune} (a), multimodal models typically consist of two encoders: an image encoder $E_i$ and a text encoder $E_t$, which extract visual features $v \in \mathbb{R}^{B\times D}$ and text features $l \in \mathbb{R}^{B\times D}$ respectively, where $B$ denotes the batch size and $D$ is the feature dimension. To effectively leverage our FFTG annotations and activate the pretrained knowledge for better forgery localization and type identification, we propose a three-branch training framework that combines unimodal and multimodal learning objectives:

\noindent\textbf{Image Feature Classification.} The visual features $v$ extracted by the image encoder $E_i$ are directly used for binary classification through a linear layer. The classification loss $L_{img}$ is defined as:
\begin{equation}
   L_{img} = -\frac{1}{B}\sum_{i=1}^B y_i\log(\text{softmax}(W_iv_i + b_i)),
\end{equation}
where $y_i \in \{0,1\}$ denotes the binary label, and $W_i, b_i$ are learnable parameters.

\noindent\textbf{Multimodal Feature Alignment.} To align visual and textual representations, we employ contrastive learning between image features $v$ and text features $l$. The alignment loss $L_{align}$ is defined as:
\begin{equation}
   L_{align} = -\frac{1}{2B}(\sum log(s(v,l) \odot I) + \sum log(s(l,v) \odot I)),
\end{equation}
where $s(\cdot,\cdot)$ denotes normalized cosine similarity and $I$ is the identity matrix.

\noindent\textbf{Multimodal Feature Classification.} We fuse visual and textual features through cross-attention and feed the fused features into a classification head. The fusion classification loss $L_{fusion}$ is:
\begin{equation}
   L_{fusion} = -\frac{1}{B}\sum_{i=1}^B y_i\log(\text{softmax}(W_f(v_i \otimes l_i) + b_f)),
\end{equation}
where $\otimes$ denotes cross-attention fusion, and $W_f, b_f$ are learnable parameters.

The overall loss function is:
\begin{equation}
   L = L_{img} + L_{align} + L_{fusion}.
\end{equation}

\subsection{Finetune Multimodol Large Language Model}

Recent advances in multimodal large language models (MLLMs) have demonstrated impressive capabilities in visual understanding and natural language reasoning. 
In addition to training visual encoders, we explore utilizing FFTG annotations to enhance the forgery detection capabilities of MLLM (e.g., LLaVA). These models typically consist of three components: a vision encoder, an alignment projector, and a large language model (LLM). In our approach, we leverage the pre-trained vision encoder from our previous step and focus on fine-tuning the alignment projector and LLM components, as shown in Figure~\ref{fig:finetune} (b).

To evaluate the model's performance, we design a straightforward yet effective prompt template: \textit{``Do you think this image is of a real face or a fake one?"} followed by \textit{``Please provide your reasons."}. This two-part prompt structure encourages the model to not only make binary decisions but also provide interpretable explanations for its judgments, enabling us to assess both the accuracy and reasoning capabilities of the fine-tuned model.



\section{Experiment}
\label{sec:experiment}

\subsection{Experimental Setting}

\begin{table*}[!h]
    \renewcommand\arraystretch{1.1}
    \centering
    \resizebox{0.95\textwidth}{!}{
    \begin{tabular}{l|ccc|cc|cccc}
    \hline
    \multirow{2}*{Method} & \multicolumn{3}{c|}{{Annotation Evaluation}}& \multicolumn{2}{c|}{CLIP Evaluation} & \multicolumn{4}{c}{MLLM Evaluation} \\
    \cmidrule{2-10}


    & Precision& Recall &F1& AVG-AUC& AVG-EER&  FFpp-ACC&CDF-ACC&Precision& Recall \\
    \hline
    w/o Text&-&-&-&84.36&20.64&50.13&65.30&10.41&8.10\\
    DD-VQA (Human)&62.46&51.52&52.06&88.25&18.04&73.54&65.60&62.94&53.62\\
    GPT-4o-mini&61.27&44.00&47.18&87.56&19.21&94.84&73.98&58.26&41.85\\
    FFTG&\textbf{89.48}&\textbf{57.12}&\textbf{64.96}&\textbf{89.08}&\textbf{17.61}&\textbf{95.84}&\textbf{75.00}&\textbf{88.07}&\textbf{55.30}\\
    \hline
    
    \hline

    \end{tabular}
    }
    \caption{Comparison of different annotation approaches. We report precision, recall and F1-score for annotation quality evaluation, AUC and EER for CLIP-based forgery detection and classification accuracy (ACC) and explanation quality (Precision/Recall) for mLLM evaluation on FFpp and Celeb-DF (CDF) datasets.
    }
    \label{table:1}
    \end{table*}

\noindent\textbf{Dataset.}
We conduct experiments on five challenging datasets: FaceForensics++~\cite{rossler2019faceforensics++}, DFDC-P~\cite{dolhansky2020deepfake}, DFD, Celeb-DF~\cite{li2019celeb}, and Wild-Deepfake~\cite{zi2020wilddeepfake}. FF++ provides paired real-fake data for generating forgery masks, while others offer diverse forgery types and scenarios. Face detection is performed using DSFD~\cite{li2019dsfd}. Detailed dataset descriptions are provided in the supplementary material.

\noindent\textbf{Annotation details.}
We use the open-source DLIB algorithm as the face landmark detector. For the forgery type decision, the threshold of mean and variance is $1.0$ and $0.5$. 
For the blur, the threshold is set to $100$. If the difference of SSIM is larger than $0.97$, we determine the forgery part is structure abnormal. The texture abnormal threshold is set to $0.7$. The blending ratio $\alpha$ is set to $0.9$. For generating refined annotations, we utilize GPT-4o-mini as our multimodal language model annotator. To create a diverse yet manageable dataset from FaceForensics++, we sample 3 frames at regular intervals from each video. During training, we use the temporally closest annotated frame as the ground truth label for intermediate frames.


\begin{table*}[!h]
    \renewcommand\arraystretch{1.1}
    \centering
    \resizebox{0.95\textwidth}{!}{
    \begin{tabular}{l|cc|cccccccc}
    \hline
    \multirow{2}*{Method} & \multicolumn{2}{c|}{\textit{FF++}}&\multicolumn{2}{c}{DFD} & \multicolumn{2}{c}{DFDC-P} & \multicolumn{2}{c}{Wild Deepfake} &\multicolumn{2}{c}{Celeb-DF}\\
    \cmidrule{2-3}
    \cmidrule{3-11}

    & AUC& EER &AUC& EER& AUC& EER& AUC& EER& AUC& EER\\
    \hline
    Xception~\cite{chollet2017xception}& 99.09&3.77 &87.86& 21.04& 69.80& 35.41&66.17 &40.14 & 65.27& 38.77\\
    EN-b4~\cite{tan2019efficientnet}   &99.22 &3.36& 87.37      & 21.99      & 70.12       & 34.54      & 61.04           & 45.34           & 68.52         & 35.61        \\
    Face X-ray~\cite{tan2019efficientnet}& 87.40&-& 85.60&-& 70.00&      -      &       -          &         -        & 74.20          &       -       \\

    F3-Net~\cite{qian2020thinking}          & 98.10& 3.58 & 86.10     & 26.17       &     72.88       &      33.38      & 67.71          &      40.17           & 71.21        &     34.03         \\
    MAT~\cite{zhao2021multi} & 99.27&3.35& 87.58      & 21.73      & 67.34       & 38.31      & 70.15           & 36.53           & 70.65         & 35.83        \\
    GFF~\cite{luo2021generalizing}             & 98.36& 3.85 & 85.51      & 25.64      & 71.58       & 34.77      & 66.51           & 41.52            & 75.31         & 32.48        \\
    LTW~\cite{sun2021domain}            & 99.17& 3.32  & 88.56      & 20.57      & 74.58       & 33.81      & 67.12           & 39.22           & 77.14         & 29.34         \\
    
    LRL~\cite{chen2021local}    & \textbf{99.46}&\textbf{3.01}& 89.24      & 20.32      & 76.53       & 32.41      & 68.76           & 37.50              & 78.26         & 29.67        \\
    
    DCL~\cite{sun2021dual}              & 99.30&3.26& 91.66      & 16.63      & 76.71       & 31.97      & 71.14         & 36.17           & 82.30       & 26.53   \\
    PCL+I2G~\cite{zhao2021learning}& 99.11&-& -      & -      & -       & -      & -         & -           & 81.80       & -   \\
    SBI~\cite{shiohara2022detecting}    & 88.33&20.47&88.13&17.25&76.53&30.22&68.22&38.11&80.76&26.97 \\
    UIA-ViT~\cite{zhuang2022uia} &-&-&94.68&-&75.80&-&-&-&82.41&- \\
    RECCE~\cite{cao2022end} &99.32&3.38&89.91&19.95&75.88&32.41&67.93&39.82&70.50&35.34 \\
    UCF~\cite{yan2023ucf} &97.05&-&80.74&-&75.94&-&-&-&75.27&- \\
    CLIP~\cite{radford2021learning}& 99.09&3.16& 89.03      & 17.13      & 78.83       & 28.95      & 77.71           & 30.38           & 77.16          & 29.30       \\

    \hline

    Ours              & 99.16&3.11& \textbf{94.81}      & \textbf{15.22}      & \textbf{83.21}       & \textbf{22.43}      & \textbf{85.10}           & \textbf{23.65}           & \textbf{83.15}          & \textbf{23.66}       \\


    \hline

    \end{tabular}
    }
    \caption{\textbf{Frame-level} cross-database evaluation from FF++(HQ) to DFD, DFDC-P, Wild Deepfake and Celeb-DF in terms of AUC and EER. The FF++ belongs to the intra-domain results while others represent the unseen-domain.
    }
    \label{table:clip}
    \end{table*}

\noindent\textbf{Training details.}
For multimodal model fine-tuning, we use CLIP with ViT-L as the image encoder. Input images are resized to $224\times 224$ pixels. The model is optimized using Adam optimizer with a learning rate of $1e-6$ and batch size of $32$. 
For MLLM fine-tuning, we use LLaVA 1.5-7b~\cite{llava-1.5} as our fundation model.
we set the learning rate to $2e-5$, batch size to $8$, gradient accumulation step to 1, and train for $3$ epochs.  All experiments are implemented in PyTorch and conducted on $4\times$ NVIDIA A100 GPUs.

\subsection{Experimental Results on FFTG}
To evaluate the quality and effectiveness of our FFTG annotations, we compare against three baseline approaches. The first baseline (\textit{w/o text}) trains the model without any textual annotations, serving as a unimodal baseline. The second baseline uses human-annotated text from DD-VQA~\cite{zhang2024common}, representing the traditional manual annotation approach. The third baseline employs GPT-4o-mini directly for annotation without our raw description guidance, demonstrating the impact of our structured prompting strategy. The experimental results are shown in Table \ref{table:1}.

We conduct comprehensive evaluations across three dimensions: 

(1) \textit{Annotation Evaluation}: Using forgery masks as ground truth, we evaluate whether generated annotations correctly identify manipulated regions (mouth, nose, eyes, face) by checking for exact terms or synonyms, measured by precision, recall, and F1-score.

(2) \textit{CLIP Evaluation}: We evaluate the classification performance using AUC and EER metrics from the Image Feature Classification branch output. We report the average metrics across five benchmark datasets to evaluate forgery detection performance.

(3) \textit{MLLM Evaluation}: We evaluate MLLM on both classification and explanation. For classification, we compute accuracy by matching the occurrence of ``real" or ``fake" in the model's response with ground truth labels. For explanation quality, we assess the accuracy of identified forgery regions following the same protocol as Annotation Evaluation.

\noindent\textbf{Annotation Evaluation.}
As shown in Table~\ref{table:1}, our FFTG significantly outperforms existing annotation methods in identifying forgery regions. FFTG achieves 89.48\% precision and 57.12\% recall, surpassing human annotations (DD-VQA) by considerable margins (27.02\% and 5.60\% respectively). Compared to direct GPT-4o mini annotation without guidance, FFTG improves precision by 28.21\% and recall by 17.12\%, resulting in a substantially higher F1-score (64.96\% vs 47.18\%). These results demonstrate that our mask-guided annotation pipeline with structured prompting effectively reduces hallucination and generates more accurate region identifications than both human annotations and direct large model outputs.

\noindent\textbf{CLIP Evaluation.}
The CLIP evaluation results demonstrate the effectiveness of incorporating textual modality and our training framework. The baseline method (w/o text), which relies solely on image features for binary classification, achieves an average AUC of 84.36\% and EER of 20.64\%. All methods with textual annotations outperform this unimodal baseline, validating the benefit of leveraging language modality to activate CLIP's pretrained knowledge. Among them, our FFTG achieves the best performance with 89.08\% AUC and 17.61\% EER, surpassing both human annotations (DD-VQA) and direct GPT-4o mini outputs by significant margins. This improvement demonstrates that high-quality text annotations, combined with our three-branch training framework, can effectively leverage the semantic knowledge embedded in pretrained CLIP model and enhance the model's forgery detection capabilities.

\begin{table}[!t]
   \centering
   \renewcommand\arraystretch{1.2}
   \resizebox{1.0\columnwidth}{!}{
   \begin{tabular}{c|c|c|c|c|c}
       \hline
       
        \multirow{2}*{Siginal}&\multirow{2}*{Method} &\multicolumn{2}{c|}{Celeb-DF}&\multicolumn{2}{c}{DFDC-P}\\
       \cline{3-6}

       && AUC &EER& AUC &EER\\
       \hline
       \multirow{1}*{Mask}&Decoder&77.70&29.56&78.51&29.59\\
       \hline
       \multirow{3}*{Digital}&Region&79.73&29.24&78.59&29.07\\
       &Type &78.45&29.95&78.17&29.92\\
       &Both &77.18&30.47&77.54&31.03\\
       \hline
       
       \multirow{4}*{Text}&Region&{81.53}&{25.11}&{80.17}&{26.35}\\
       
       &Type&{80.40}&{27.11}&{78.25}&{28.19}\\

       &Both (Raw)&{82.15}&{24.58}&{81.55}&{24.12}\\

       &Ours&\textbf{83.15}&\textbf{23.66}&\textbf{83.21}&\textbf{22.43}\\

       \hline
   \end{tabular}
   }
   \caption{Ablation study on different supervisory signals. }
   \label{table:ablation1}
\end{table}

\noindent\textbf{MLLM Evaluation.}
For MLLM evaluation, we assess both the classification accuracy and explanation quality of fine-tuned models. In terms of classification, our FFTG-enhanced model achieves the highest accuracy of 95.84\% on FFpp (intra-domain) and 75.00\% on Celeb-DF (cross-domain), significantly outperforming the baseline without text (50.13\% and 65.30\%). Notably, while DD-VQA annotations show moderate improvement (73.54\% on FFpp), and direct GPT-4o-mini annotations achieve competitive accuracy (94.84\% on FFpp), our method consistently performs better across different datasets, demonstrating more robust generalization.

For explanation quality, FFTG generates more accurate forgery region identifications with 88.07\% precision and 55.30\% recall, substantially surpassing both human annotations and direct GPT-4o-mini outputs . These results validate that our structured prompting strategy not only improves the model's classification capability but also enhances its ability to provide accurate and reliable explanations for its decisions, which is crucial for practical applications requiring interpretable outputs.

\subsection{Comparison with State-of-the-Art Methods}

\noindent\textbf{Cross-dataset evaluation.}
To evaluate the generalization capability of our fine-tuned CLIP model, we conduct extensive experiments across multiple deepfake datasets. Following standard protocol, we train our model on the high-quality version of FF++ and test on other challenging datasets that exhibit significant domain gaps in terms of forgery types, human identities, video backgrounds, and image quality. 

The quantitative results are shown in Table~\ref{table:clip}. Our method achieves consistent improvements across all unseen datasets. Specifically, on DFDC-P, our method achieves 83.21\% AUC, surpassing the recent transformer-based method UIA-ViT (75.80\%) by a significant margin of 7.41\%. On the challenging Wild Deepfake dataset, our approach reaches 85.10\% AUC, outperforming DCL by nearly 14\%. For Celeb-DF, we achieved 83.15\% AUC, demonstrating superior performance compared to both traditional methods and recent advances like PCL+I2G (81.80\%). These substantial improvements can be attributed to two key factors: 1) the high-quality text annotations from FFTG that help activate CLIP's pretrained knowledge, and 2) our effective three-branch training framework that facilitates both unimodal and multimodal feature learning.

\subsection{Ablation Study}

\begin{table}[!t]
    \centering
    \renewcommand\arraystretch{1.2}
    \resizebox{\columnwidth}{!}{
    \begin{tabular}{c|c|cc|cc}
        \hline
        \multirow{2}*{Alignment}&\multirow{2}*{Multimodal} & \multicolumn{2}{c|}{Celeb-DF}&
        \multicolumn{2}{c}{Wild Deepfake}  \\

        \cline{3-6}
        && AUC& EER &AUC& EER \\
        \hline
        $\times$&$\times$&77.16&29.30&77.71&30.38\\
        $\checkmark$&$\times$&82.19&24.76&82.25&26.77\\
        $\times$&$\checkmark$&81.66&24.31&80.35&28.13\\
        
        $\checkmark$&$\checkmark$&\textbf{83.15}&\textbf{23.66}&\textbf{85.10}&\textbf{23.65}\\
        \hline
    \end{tabular}
    }
    \vspace{-2pt}
    \caption{Ablation study on the impact of different components in terms of AUC and EER. `Alignment' indicates the Multimodal Feature Alignment ($L_{align}$). `Multimodal' signifies the Multimodal Feature Classification ($L_{fusion}$).
    }
    \label{table:ab2}
\end{table}

\begin{figure*}[!t]
    \begin{center}
    
       \includegraphics[width=1.0\linewidth]{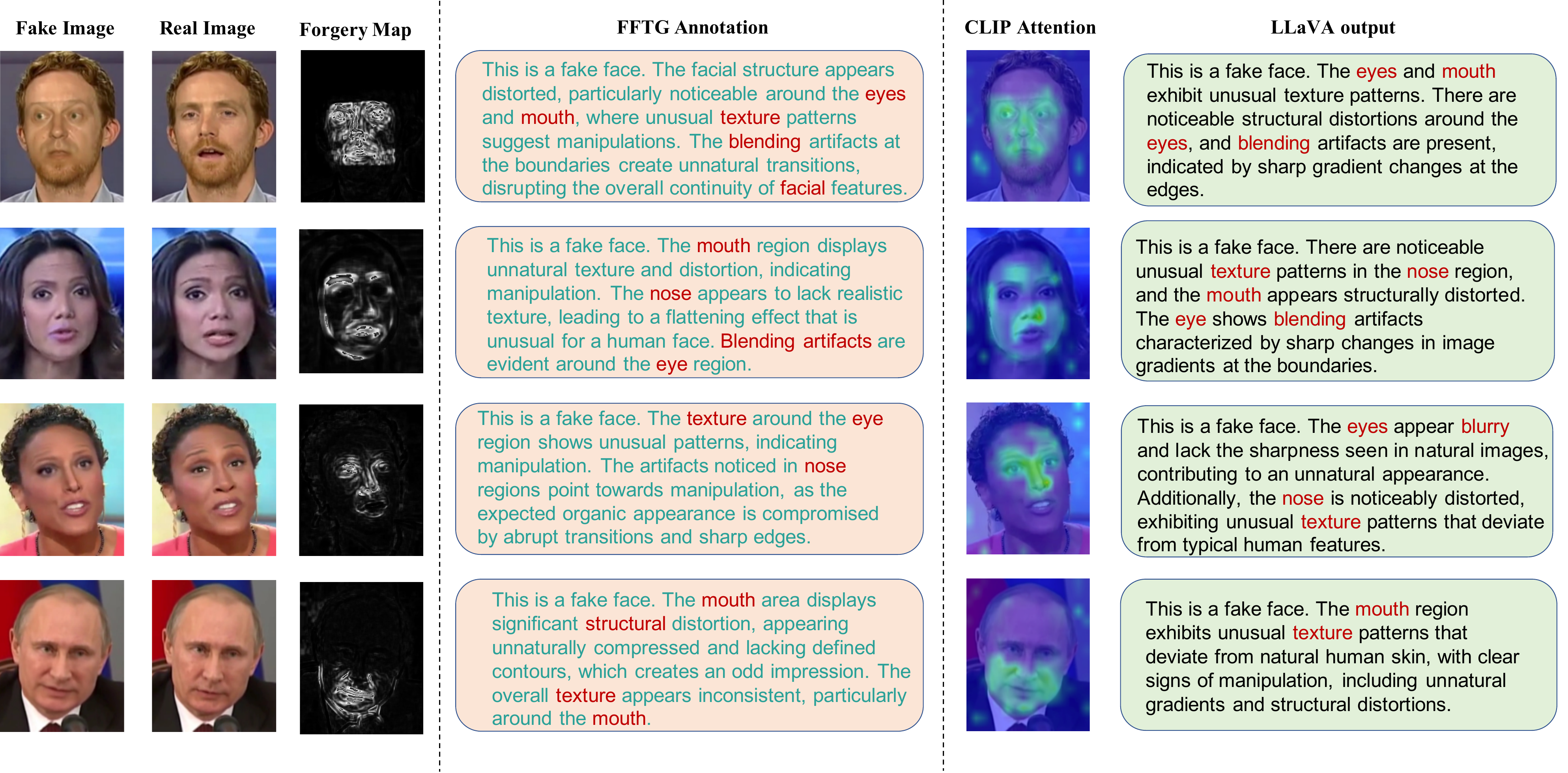}
    \end{center}
       \caption{ Visualization of FFTG annotation pipeline and model inference results. For each example, we show the fake-real image pair, forgery mask, FFTG's annotation, CLIP attention map, and LLaVA's output. FFTG annotations align well with forgery masks and guide both CLIP and LLaVA to focus on genuine manipulation regions.
       }
    \label{fig:vis1}
\end{figure*}

\noindent\textbf{Impact of language information.}
To investigate the effectiveness of different supervisory signals, we compare three approaches: mask-based, digital label-based, and text-based supervision. For mask-based supervision, we employ a decoder to regress forgery masks. For digital supervision, we experiment with region-only, type-only, and both labels as classification targets. As shown in Table~\ref{table:ablation1}, mask-based supervision achieves limited performance (77.70\% AUC on Celeb-DF), likely due to overfitting to low-level features. Digital supervision performs slightly better, with region-based classification reaching 79.73\% AUC.

Our text-based approach significantly outperforms both alternatives, achieving 83.15\% AUC on Celeb-DF and 83.21\% on DFDC-P. This improvement can be attributed to two factors: 1) the rich semantic information captured by textual descriptions compared to binary or categorical labels, and 2) our FFTG pipeline that generates accurate and diverse annotations. The results also show that using both region and type information (Raw Annotation) performs better than using either alone, demonstrating the benefit of comprehensive text descriptions in guiding the model's learning process.


\noindent\textbf{Impact of different components.}
To analyze the effectiveness of our three-branch training framework for fintuning multimodal model, we conduct ablation studies on two key components: Multimodal Feature Alignment and Multimodal Feature Classification. As shown in Table~\ref{table:ab2}, using only Image Feature Classification achieves baseline performance (77.16\% AUC on Celeb-DF). Adding Multimodal Feature Alignment improves the AUC by 4.5\%, demonstrating the benefit of aligning visual and textual representations. Incorporating Multimodal Feature Classification further boosts performance by leveraging cross-attention fusion. The full model combining all three components achieves the best results (83.15\% AUC on Celeb-DF and 85.10\% AUC on Wild Deepfake), indicating that the different components complement each other in learning discriminative features for forgery detection.

\subsection{Visualizations}



Figure~\ref{fig:vis1} showcases our analysis pipeline on FFpp test set examples. FFTG annotations demonstrate high accuracy in identifying manipulated regions and describing forgery types. For instance, in the first example, our annotation captures both eyes and mouth regions distortion along with blending artifacts at boundaries, precisely matching the forgery mask. In the second case, the annotation identifies mouth region's unnatural texture and nose's unrealistic appearance, while also noting blending artifacts around eye regions. The third example shows detailed description of texture abnormalities around eyes and nose, while the last example accurately captures the mouth region's structural distortion and texture inconsistencies.

The attention maps from our fine-tuned CLIP model exhibit strong alignment with forgery masks, particularly evident in high-attention areas matching manipulated regions. For example, in the second row, CLIP's attention clearly highlights both the nose and mouth regions identified in the forgery mask. Similarly, LLaVA outputs demonstrate enhanced detection capabilities after fine-tuning, providing precise and consistent explanations. In the third example, LLaVA correctly identifies both the "blurry" appearance of eyes and the distorted nose with unusual texture patterns, showing strong correlation with FFTG annotations and forgery masks.

Due to space limitations, additional visualizations including baseline comparisons and dialogue examples are provided in supplementary materials.
\section{Conclusion}
\label{sec:conclusion}

In this paper, we analyze the limitations of existing text annotation approaches and present Face Forgery Text Generator (FFTG), a novel annotation pipeline that combines mask-guided analysis with structured prompting strategies to generate accurate and interpretable text descriptions for face forgery detection. Our extensive experiments demonstrate that FFTG effectively addresses the hallucination issues in existing annotation methods, achieving higher accuracy in region identification and leading to substantial improvements when fine-tuning both CLIP and MLLM. These results validate the importance of high-quality text annotations in enhancing both the generalization and interpretability of forgery detection systems, providing a promising direction for future research in multimodal forensics tasks.

{
    \small
    \bibliographystyle{ieeenat_fullname}
    \bibliography{main}
}
\clearpage
\setcounter{page}{1}
\maketitlesupplementary
\appendix

\section*{Overview of Supplementary Materials}
This supplementary material provides additional details and experimental results to support our main paper. It is organized as follows:
\begin{itemize}
    \item Section A details the FFTG algorithm's forgery type decision criteria and procedures.
    \item Section B presents additional experimental results on cross-manipulation and multi-source evaluation.
    \item Section C describes the dataset details and training protocols.
    \item Section D provides comprehensive visualizations including attention maps, annotation comparisons, and LLaVA responses.
    \item Section E explains the prompt design and implementation details.
\end{itemize}

\section{Details of FFTG}
This section mainly introduces the details of the forgery type decision in the FFTG algorithm. 
\label{sec:A}

\noindent\textbf{Color Difference.}
This phenomenon occurs in the face swap when the color of the source and target face has a drastic difference. Inspired by the color transfer~\cite{reinhard2001color},
we leverage the distance of the average channel-wise mean and variance of the real and fake regions in the $Lab$ color space to determine whether there exists a color difference. The $Lab$ color space minimizes correlation between channels, which helps reduce the impact of changes in a certain channel on the overall color.
The pseudocode is shown in Alg.~\ref{alg:cd}, $split$ represents dividing the channel of the image, $Lab$ denotes converting the RGB color space into $Lab$ space.

\begin{algorithm}[!h]
    
    \caption{Color Difference Decision}
    \begin{algorithmic}[1]
        \Require{Real image selected region $R_s(i_r)$, fake image selected region $R_s(i_f)$, mean threshold $\theta_c^m$, standard deviation threshold$\theta_c^s$}
        \State $R_s(i_r)^{'},R_s(i_f)^{'} = Lab(R_s(i_r)),Lab(R_s(i_f))$
        \State $L_r,a_r,b_r = split(R_s(i_r)^{'})$
        \State $L_f,a_f,b_f = split(R_s(i_f)^{'})$
        \State $L^m = ||mean(L_r) - mean(L_f)||_2$
        \State $a^m = ||mean(a_r) - mean(a_f)||_2$
        \State $b^m = ||mean(b_r) - mean(b_f)||_2$
        \State $L^s = ||std(L_r) - std(L_f)||_2$
        \State $a^s = ||std(a_r) - std(a_f)||_2$
        \State $b^s = ||std(b_r) - std(b_f)||_2$
        \State m = ($L^m$ + $a^m$ + $b^m$) / 3
        \State s = ($L^s$ + $a^s$ + $b^s$) / 3
        \If {m $> \theta_c^m$ and s $> \theta_c^s$}
            \State \textbf{Return} True
        \Else
            \State \textbf{Return} False
        \EndIf
    \end{algorithmic}
    \label{alg:cd}
    
\end{algorithm}

\noindent\textbf{Blur.}
There exists local blurring in forgery faces due to the instability of the generated model or blending operation. To quantify such phenomena, we make use of the Laplacian image, which can reflect the sharpness of image edges.
Specifically, as shown in Alg.~\ref{alg:blur}, we compute the variance of the real and fake images of the selected region after the Laplacian operator, and if the value of the real is larger than the fake one and their difference is greater than a certain threshold, we define this part as blurred. The $Laplacian(.)$ represents the Laplacian operator, $var(.)$ means calculating the variance of the input image.

\noindent\textbf{Structure Abnormal.}
We observed that compared with normal faces, some organs of fake faces will be obviously deformed. To metric such structure deformable, we use the Structural Similarity (SSIM) index difference between real and fake images of the selected region $R_s$ to decide whether the chosen region has a structure abnormal or not, which details in Alg.~\ref{alg:sab}.

\begin{algorithm}[!h]
    
    \caption{Blur Decision}
    \begin{algorithmic}[1]
        \Require{Real image selected region $R_s(i_r)$, fake image selected region $R_s(i_f)$, variance threshold $\theta_b^v$}
        \State $r\_var = var(Laplacian(R_s(i_r)))$
        \State $f\_var = var(Laplacian(R_s(i_f)))$

        \If {$r\_var > f\_var$ and ($r\_var - f\_var) > \theta_b^v$}
            \State \textbf{Return} True
        \Else
            \State \textbf{Return} False
        \EndIf
    \end{algorithmic}
    \label{alg:blur}
    
\end{algorithm}

\noindent\textbf{Texture Abnormal.}
It has been proved that the generator typically correlates the values of nearby pixels and cannot generate as strong texture contrast as real data~\cite{liu2020global}, leading to texture differences in some forgery regions. Similar to the Gram-Net~\cite{liu2020global}, we leverage a texture analysis tool--the contrast of Gray-Level Co-occurrence Matrix (GLCM)~\cite{haralick1973textural}, formed as $C_d$.
Larger $C_d$ reflects stronger texture contrast, sharper and clearer visual effects. Inversely, a low value $C_d$ means the texture is blurred and unclear.
We define a forgery region as texture abnormal when the $C_d$ of the real is larger than the fake one beyond the threshold. The algorithm is shown in Alg.~\ref{alg:tab}, where $GLCM$ represents the average Gray-Level Co-occurrence Matrix of the input from right, down, left, and upper four orthogonal directions.

\begin{algorithm}[!h]
    
    \caption{Structure Abnormal Decision}
    \begin{algorithmic}[1]
        \Require{Real image selected region $R_s(i_r)$, fake image selected region $R_s(i_f)$, ssim threshold $\theta_s$}
        \State $s = ssim(R_s(i_r),R_s(i_f))$
        \If {$s < \theta_s$}
            \State \textbf{Return} True
        \Else
            \State \textbf{Return} False
        \EndIf
    \end{algorithmic}
    \label{alg:sab}
    
\end{algorithm}

\begin{algorithm}[!h]
    
    \caption{Texture Abnormal Decision}
    \begin{algorithmic}[1]
        \Require{Real image selected region $R_s(i_r)$, fake image selected region $R_s(i_f)$, contrast threshold $\theta_t$}
        \Ensure{ $N=256 \times 256$}
        \State $P_r = GLCM(R_s(i_r))$
        \State $P_f = GLCM(R_s(i_f))$
        \State $C_d^r = \frac{1}{N}\sum_{i=0}^{255}\sum_{j=0}^{255}|i-j|^2P_r(i,j)$
        \State $C_d^f = \frac{1}{N}\sum_{i=0}^{255}\sum_{j=0}^{255}|i-j|^2P_f(i,j)$

        \If {$C_f^r > C_d^f$ and ($C_d^r - C_d^f) > \theta_t$}
            \State \textbf{Return} True
        \Else
            \State \textbf{Return} False
        \EndIf
    \end{algorithmic}
    \label{alg:tab}
    
\end{algorithm}

\noindent\textbf{Blend Boundary.}
Existing face manipulation methods often leave intrinsic cues at the blending boundaries when merging manipulated faces with original backgrounds. As detailed in Alg.~\ref{alg:blend}, we first extract inner ($I_{inner}$) and outer ($I_{outer}$) boundary regions around the manipulation mask to analyze the transition area where artifacts typically occur. We then analyze three key characteristics: gradient discontinuity assessed by comparing mean gradient magnitudes between inner and outer regions using Sobel operators to identify abrupt changes in intensity transitions, edge artifacts detected through Canny detection on the combined boundary region where manipulation often creates abnormal edge densities and patterns at the interface between real and fake regions, and frequency domain abnormalities examined by analyzing the ratio of high to low frequency components in the DCT transform of the boundary area, as blending operations typically introduce unnatural high-frequency patterns that differ from smooth transitions in natural images. By analyzing the combined boundary region rather than separate inner and outer regions for edge and frequency analysis, we can better capture the complete transition patterns and avoid missing artifacts that occur exactly at the boundary interface. The detection combines these multiple evidence sources to ensure reliability, requiring at least two metrics to exceed their thresholds before classifying a region as containing significant blending artifacts, thus reducing false positives while maintaining sensitivity to various types of blending anomalies.

\begin{algorithm}[!h]
\caption{Blend Boundary Decision}
\begin{algorithmic}[1]
\Require{Image region $I$, mask $M$, threshold set ${\theta_g, \theta_e, \theta_f}$}
\State // Get boundary regions
\State $I_{inner}, I_{outer} = GetBoundaryRegion(M)$
\State // Check gradient discontinuity
\State $g_x = Sobel(I, x)$, $g_y = Sobel(I, y)$
\State $g_{mag} = \sqrt{g_x^2 + g_y^2}$
\State $s_g = |mean(g_{mag}[I_{inner}]) - mean(g_{mag}[I_{outer}])|$
\State // Check edge artifacts
\State $E = Canny(I)$
\State $s_e = sum(E * (I_{inner} + I_{outer})) / sum(I_{inner} + I_{outer})$
\State // Check frequency patterns
\State $F = DCT(I * (I_{inner} + I_{outer}))$
\State $s_f = sum(|F_{high}|) / sum(|F_{low}|)$
\State // Count evidence
\State $evidence = 0$
\If{$s_g > \theta_g$} $evidence += 1$ \EndIf
\If{$s_e > \theta_e$} $evidence += 1$ \EndIf
\If{$s_f > \theta_f$} $evidence += 1$ \EndIf
\State \Return $evidence \geq 2$
\end{algorithmic}
\label{alg:blend}
\end{algorithm}

\section{Additional Experimental Results}
\label{sec:exp}
\subsection{Cross-manipulation evaluation}


To further validate the generalization capability of our FFTG-enhanced CLIP model, we conduct cross-manipulation experiments using the high-quality version of FF++ dataset. We train our model on one manipulation method and evaluate it on all four methods (DeepFakes (DF), Face2Face (F2F), FaceSwap (FS), and NeuralTextures (NT)) to assess detection performance on unseen manipulation types.
As shown in Table ~\ref{table:2}, we compare our approach with three recent state-of-the-art methods: Multi-attentional (MAT), GFF, and DCL. The diagonal values represent intra-domain performance, while off-diagonal values indicate cross-manipulation generalization. Our method demonstrates superior performance in most scenarios, particularly in challenging cross-manipulation cases. For instance, when training on FaceSwap and testing on DeepFakes, our method achieves 87.55\% AUC, surpassing DCL by 13\%. The improvements can be attributed to the high-quality text annotations generated by FFTG and our three-branch training framework, which help the model capture manipulation patterns that are common across different forgery types.

\begin{table}[!t]
    \centering
    \renewcommand\arraystretch{1.1}
    \resizebox{1,0\columnwidth}{!}{
    \begin{tabular}{c|c|cccc}
    \hline
    Train   & Method       & DF    & F2F   & FS    & NT    \\
    \hline
    \multirow{4}{*}{DF}  
    & MAT          & \textit{99.92} & 75.23 & 40.61 & 71.08 \\
    & GFF          & \textit{99.87}	&76.89	&47.21&	72.88 \\
    & DCL    & \textit{\textbf{99.98}} & 77.13 & 61.01 & 75.01 \\
    & Ours    & \textit{99.91} & \textbf{85.41} & \textbf{75.34} & \textbf{77.19} \\
    \hline
    \multirow{3}{*}{F2F} 
    & MAT          & 86.15&	\textit{99.13}&	60.14&	64.59 \\
    & GFF          & 89.23	&\textit{99.10}&	61.30&	64.77 \\
    & DCL    & 91.91  & \textit{99.21} & 59.58 & 66.67 \\
    & Ours    & \textbf{92.32} & \textit{\textbf{99.35}} & \textbf{62.19} & \textbf{67.81} \\
    \hline
    \multirow{3}{*}{FS}  
    & MAT          & 64.13	&66.39&\textit{99.67}	&	50.10 \\
    & GFF          &70.21&	68.72 &	\textit{99.85}&	49.91 \\
    & DCL    & 74.80  &   69.75&\textit{\textbf{99.90}}  &52.60  \\
    & Ours    & \textbf{87.55} & \textbf{79.13} & \textit{{99.27}} & \textbf{53.53} \\
    \hline
    \multirow{3}{*}{NT}  
    & MAT          &  87.23	&48.22&	75.33&	\textit{98.66}  \\
    & GFF          & 88.49	&49.81&	74.31&	\textit{98.77} \\
    & DCL    & 91.23 & 52.13 &79.31 & \textit{98.97}\\
    & Ours    & \textbf{93.10} & \textbf{61.55} & \textbf{83.27} & \textit{\textbf{98.98}} \\
    \hline
                         
    \end{tabular}
    }
    \caption{Cross-manipulation evaluation in terms of AUC. Diagonal results indicate the intra-domain performance.
    }
    \label{table:2}

    \end{table}

\subsection{Multi-source manipulation evaluation.}


We evaluate the model's generalization capability through multi-source manipulation experiments, where we train on three manipulation methods and test on the remaining unknown method. This challenging protocol assesses the model's ability to detect previously unseen manipulation types. The experiments are conducted on both high-quality (HQ) and low-quality (LQ) versions of FF++ dataset to comprehensively evaluate robustness across different image qualities.
As shown in Table \ref{table:4}, our method consistently outperforms existing approaches across all settings. On high-quality DeepFakes (DF-HQ), our method achieves 95.07\% accuracy, surpassing the previous state-of-the-art UIA-ViT by 4.67\%. Similar improvements are observed for Face2Face (F2F) detection, where we achieve 88.12\% accuracy on HQ data. Notably, the performance advantage is maintained in low-quality scenarios, where compression artifacts make forgery detection particularly challenging. For instance, on DF-LQ and F2F-LQ, our method achieves 86.17\% and 71.25\% accuracy respectively, significantly outperforming previous methods like DCL and EN-B4. These results demonstrate that our FFTG-enhanced approach not only excels at detecting high-quality forgeries but also maintains robust performance when dealing with compressed, low-quality images, suggesting effective learning of manipulation-specific features that persist across different image qualities.

\begin{table}[!t]
    \centering
    \renewcommand\arraystretch{1.1}
    \resizebox{1.0\linewidth}{!}{
        \scalebox{1}{
    \begin{tabular}{c|c|c|c|c}
        \hline
        \multirow{2}*{Method}& \multicolumn{1}{|l|}{DF (HQ)} & \multicolumn{1}{l|}{DF (LQ)} & \multicolumn{1}{l|}{F2F (HQ)} & \multicolumn{1}{l}{F2F (LQ)} \\
        \cline{2-5}
    & ACC                       & ACC                     & ACC            &  ACC                    \\
    \hline
    EN-B4         & 82.40                    & 67.60                     & 63.32                     & 61.41                      \\
    Focalloss        & 81.33                    & 67.47                    & 60.80                  & 61.00                        \\
    Multi-task       & 70.30                        & 66.76                        & 58.74                         & 56.50                         \\
    MLDG             & 84.21                    & 67.15                      & 63.46                      & 58.12                     \\
    LTW              & 85.60                     & 69.15                    & 65.60                      & 65.70                      \\

    DCL             & 87.70                   & 75.90                   & 68.40                    & 67.85              \\
    
    
    UIA-ViT    & 90.40                   &  -              & 86.40             & -         \\
    \hline   
    Ours             & \textbf{95.07}                    & \textbf{86.17}                   & \textbf{88.12}                    & \textbf{71.25}              \\
    \hline
    \end{tabular}
    }
    }
    \caption{Performance on multi-source manipulation evaluation, the protocols and the compaired results are from \cite{rossler2019faceforensics++}. DF means traning on the other three manipulated methods of FFpp and test on deepfakes class. The same for the others.}
    \label{table:4}
    \end{table}

\section{Dataset Details}
\subsection{Training and Test dataset.}
To evaluate the generalization of our proposed annotation, we 
conduct our experiments on several challenging datasets:
1) FaceForensics++~\cite{rossler2019faceforensics++}: a widely-used forgery dataset contains 1000 videos with four different manipulated approaches, including two deep learning based \textit{DeepFakes} and \textit{NeuralTextures} and two graphics-based methods \textit{Face2Face} and \textit{FaceSwap}. This dataset provides pairwise real and forgery data, enabling us to generate mixed forgery images with FFTG. 2) DFDC-P~\cite{dolhansky2020deepfake} dataset is a challenging dataset with $1133$ real videos and $4080$ fake videos, containing various manipulated methods and backgrounds.
3) DFD is a forgery dataset containing $363$ real videos and $3068$ fake videos, which is mostly generated by the Deepfake method. 4) Celeb-DF~\cite{li2019celeb}
is another high-quality Deepfake dataset that contains various scenarios. 5) Wild-Deepfake~\cite{zi2020wilddeepfake} is a forgery face dataset
obtained from the internet, leading to a diversified distribution of scenarios. We use DSFD~\cite{li2019dsfd} to extract faces from each video.
\begin{figure*}[!t]
    \begin{center}
    
       \includegraphics[width=0.9\linewidth]{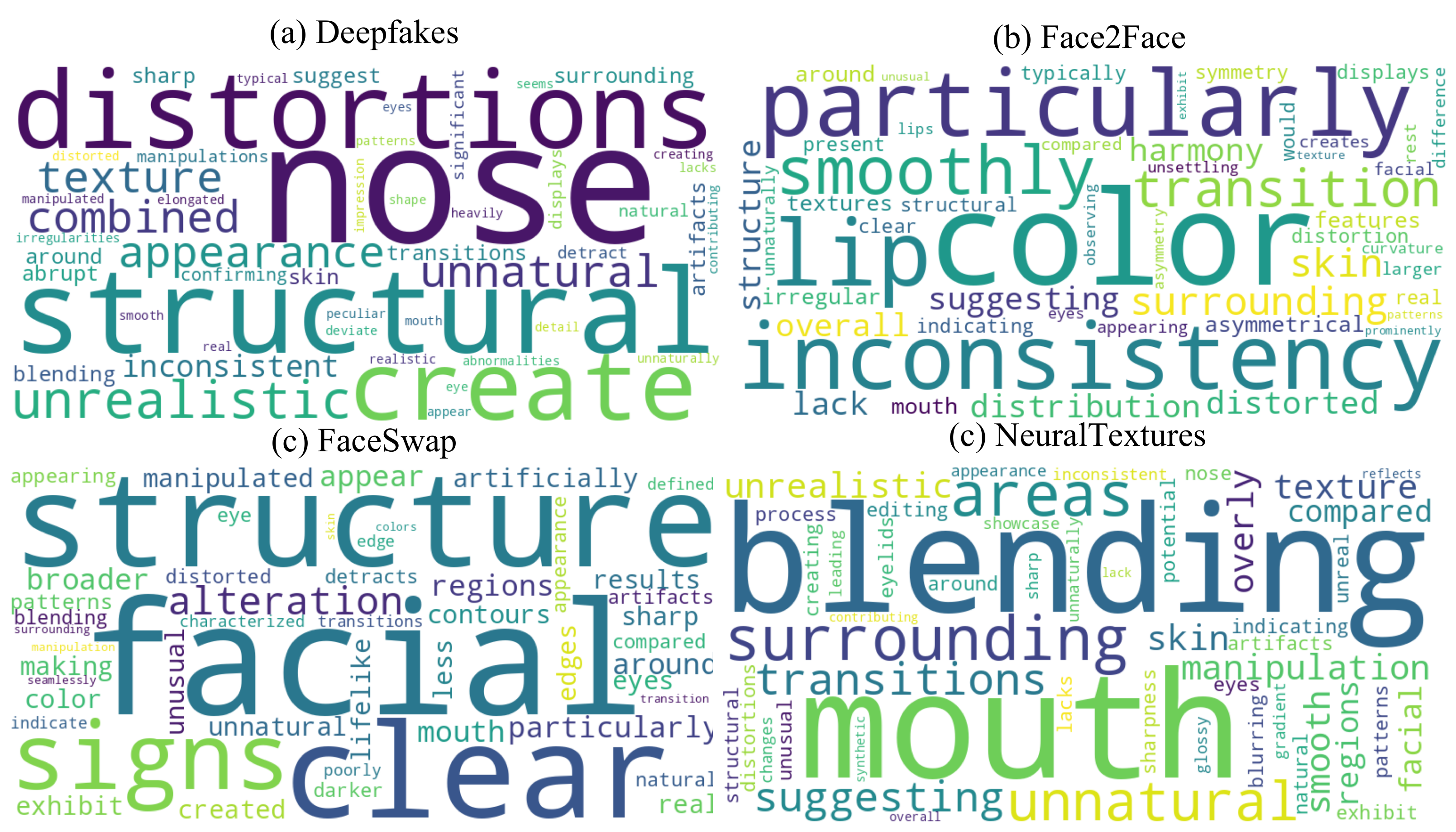}
    \end{center}
       \caption{ Word cloud comparison of FFTG annotations on FFpp dataset.
       }
    \label{fig:fftg_wcloud}
\end{figure*}

\subsection{Analysis of Text Annotations}
To better understand the characteristics of FFTG annotations across different manipulation types, we visualize their word distributions through word clouds in Figure \ref{fig:fftg_wcloud}. In Deepfakes, the annotations concentrate on structural aspects, with "distortions" and "nose" being prominent, along with texture-related descriptions, reflecting the method's tendency to create geometric inconsistencies. For Face2Face, the word cloud reveals a focus on color inconsistencies and transitions, with terms like "lipcolor" and "particularly" frequently appearing, indicating the method's impact on local appearance details. In FaceSwap cases, FFTG identifies broader structural changes, with "facial" and "structure" being dominant terms, while also capturing clear signs of alterations in face contours. The NeuralTextures annotations emphasize blending-related artifacts, with "blending" and "surrounding" appearing prominently, along with specific attention to mouth regions and transitions.
This visualization demonstrates FFTG's ability to generate precise, manipulation-specific annotations that capture the unique characteristics of each forgery type. The focused vocabulary and consistent emphasis on specific artifacts reflect the effectiveness of our mask-guided approach in identifying and describing relevant manipulation features.

    
    

    
    

\section{Additional Visuallization}

\subsection{Visualizations on FFpp dataset.}
\begin{figure}[!t]
    \begin{center}
    
      \includegraphics[width=1\linewidth]{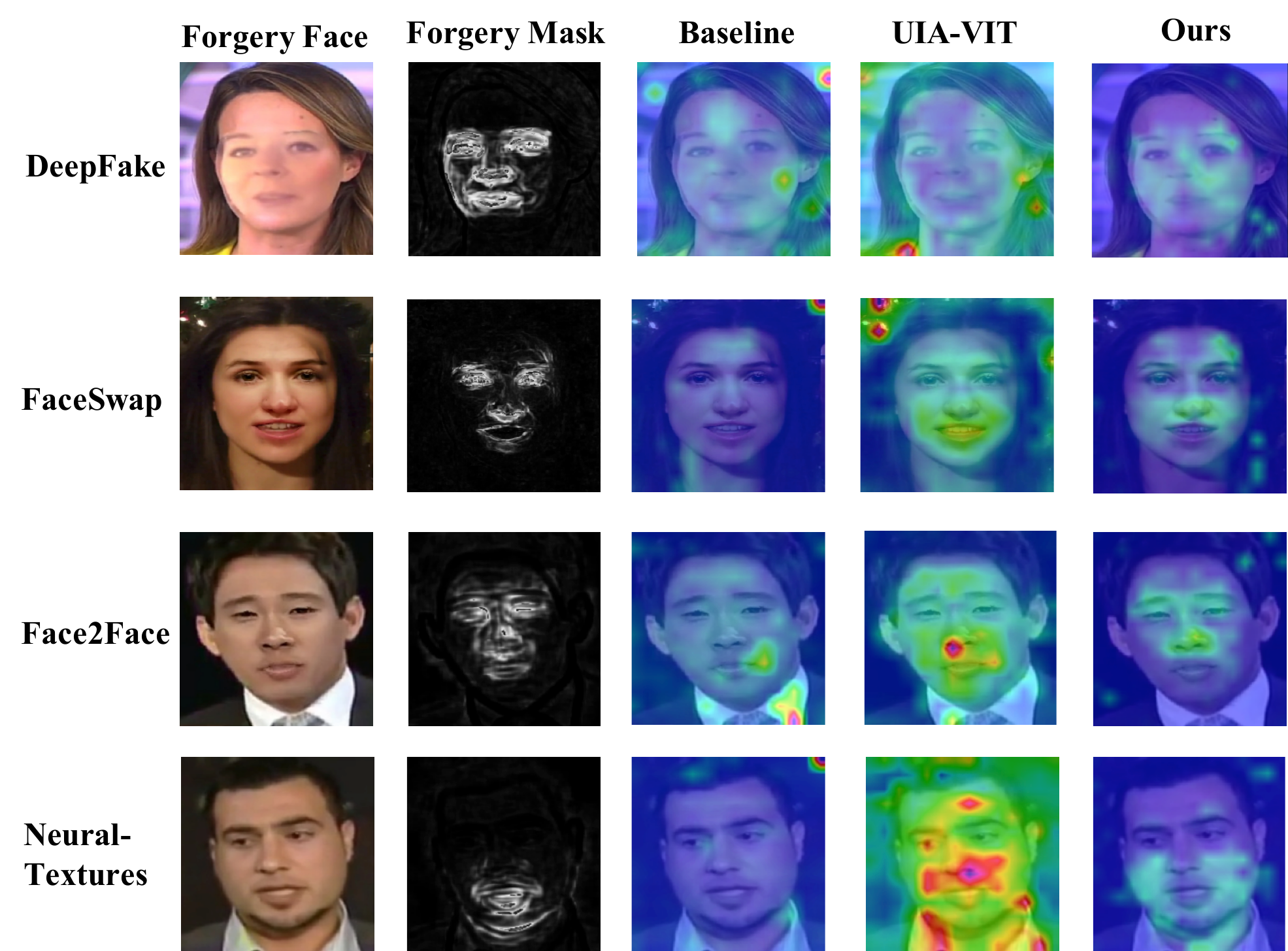}
    \end{center}
      \caption{Visualization of attention heatmap on training dataset (FFpp) of the baseline, UIA-VIT, and our proposed method. The forgery Mask represents the ground truth 
      manipulation mask generated by Eq. 1.
      }
    \label{fig:cam_mask}
\end{figure}

To further validate the interpretability of our method, we visualized the attention heatmaps across different approaches on the test set of FFpp HQ dataset, comparing our method with a baseline (binary classification with CLIP pretrained image-encoder) and the state-of-the-art UIA-VIT~\cite{zhuang2022uia}. The comparison in Figure~\ref{fig:cam_mask} spans four manipulation methods: DeepFake, FaceSwap, Face2Face and NeuralTextures, with corresponding ground truth masks serving as references for manipulation regions.
The baseline model shows diffused attention patterns that lack precise localization of manipulated regions. UIA-VIT demonstrates improved focus but still exhibits scattered attention that sometimes deviates from the actual manipulation areas. In contrast, our method achieves significantly more precise attention localization that closely aligns with the ground truth masks across all manipulation types. This is particularly evident in the NeuralTextures example, where our method accurately concentrates on the subtle mouth area manipulations while other methods show misplaced or dispersed attention. For Deepfake and FaceSwap cases, our attention maps precisely highlight the key manipulated facial regions, and in Face2Face examples, they effectively capture the structural modifications. This precise alignment between our attention maps and ground truth masks demonstrates that the fine-grained linguistic supervision from FFTG annotations effectively guides the model to focus on genuine manipulation artifacts, improving both detection accuracy and interpretability.

\subsection{Visualizations on unseen dataset.}
\begin{figure}[!t]
    \begin{center}
      \includegraphics[width=1\linewidth]{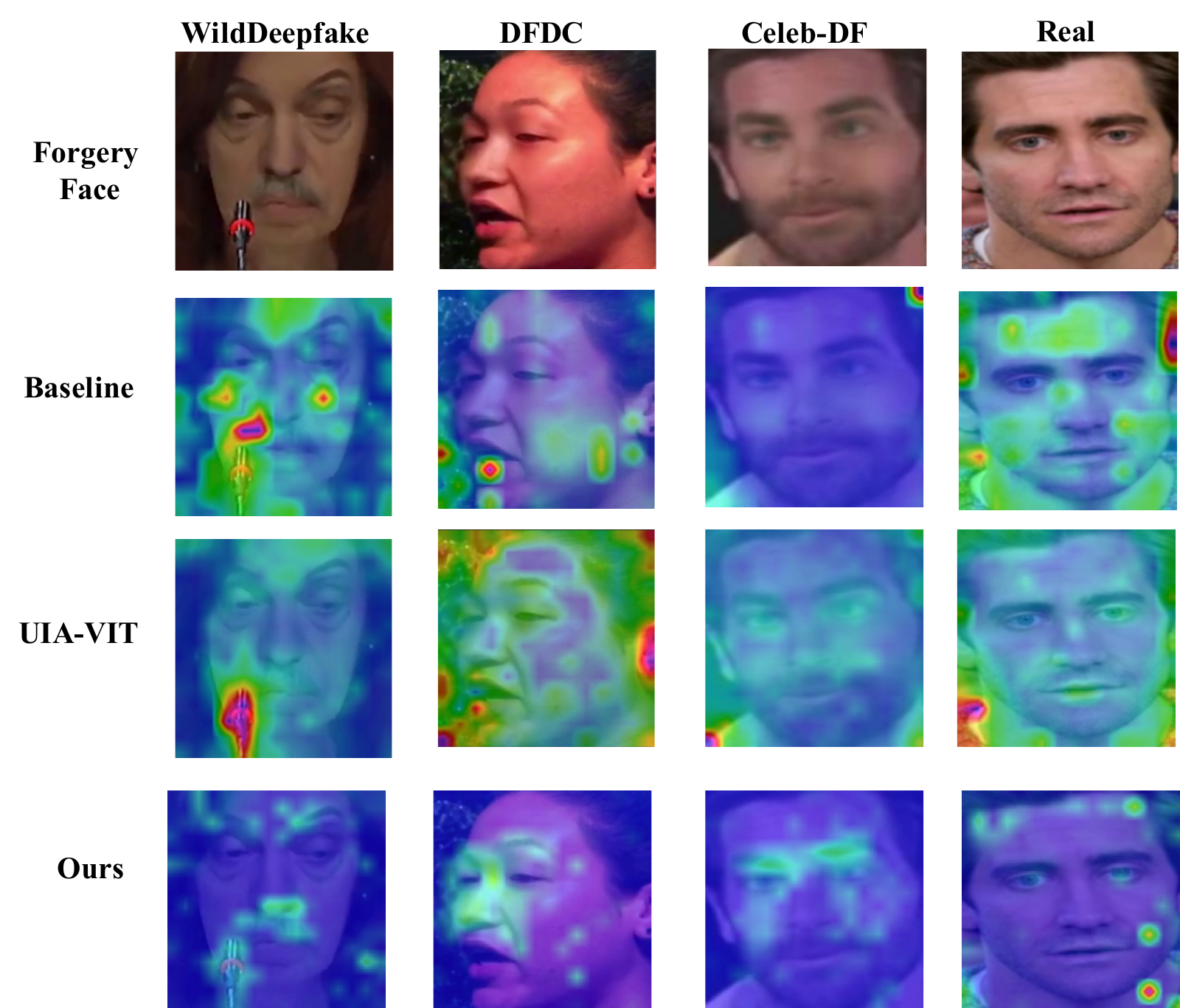}
    \end{center}
      \caption{Attention heatmap visualization of the baseline, UIA-VIT, and our proposed method on the unseen dataset. The first row represents the original images that did not appear in the training set.
      }
    \label{fig:cam_cross}
\end{figure}

We visualize attention maps from different models on various unseen datasets (WildDeepfake, DFDC, and Celeb-DF) along with real faces in Figure \ref{fig:cam_cross}. The baseline model's attention appears scattered and unfocused, with activation spread across irrelevant facial regions, indicating its limited ability to identify manipulation-specific features. UIA-ViT shows improved attention patterns with better concentration on facial components, but still exhibits some dispersion and occasionally highlights unmanipulated areas. In contrast, our method demonstrates more precise attention localization that aligns well with actual forgery regions. For instance, in WildDeepfake samples, our model precisely concentrates on the manipulated facial features while maintaining minimal activation on unmodified areas. On DFDC and Celeb-DF, it effectively captures the subtle manipulation artifacts despite their varying characteristics. When processing real faces, our model maintains clean and evenly distributed attention patterns without false activations. These visualizations confirm that our FFTG-guided approach helps the model learn more accurate and interpretable features for face forgery detection, enabling better generalization across different domains and manipulation types.

\subsection{Visualizations of Annotation}
To better understand the differences between annotation methods and demonstrate FFTG's advantages, we provide a detailed comparison of annotations generated by different approaches across four major manipulation types: Deepfakes, Face2Face, FaceSwap, and NeuralTextures. We present the manipulated image, forgery mask, real image, and corresponding annotations from human annotators, GPT-4o, DD-VQA, and our FFTG method, with key forgery-related terms highlighted in red to emphasize each method's detection focus.

\begin{figure*}[!t]
    \begin{center}
    
       \includegraphics[width=0.8\linewidth]{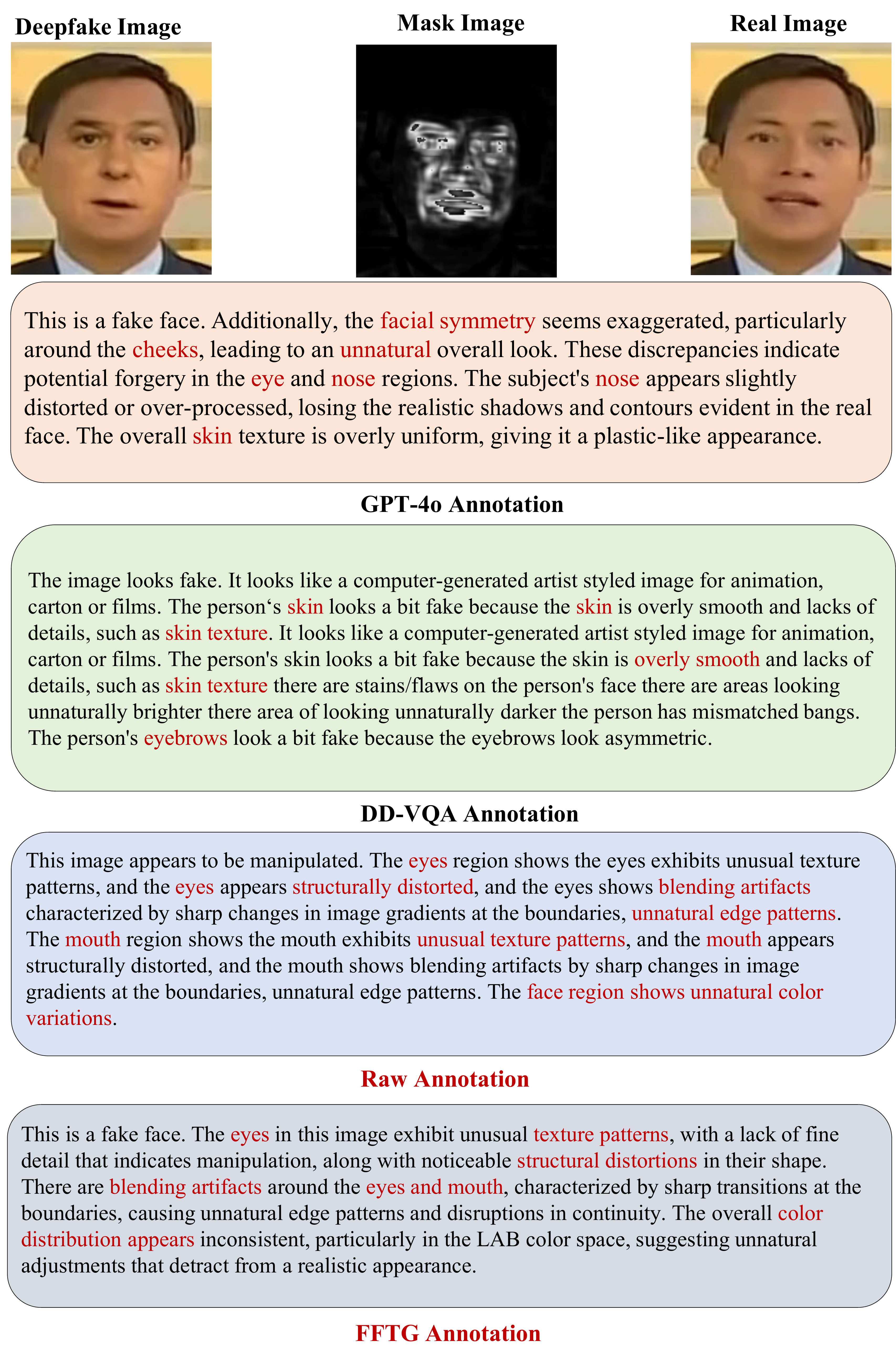}
    \end{center}
       \caption{Comparison of different annotation approaches on a Deepfake example.
       }
    \label{fig:df_anno}
\end{figure*}

As shown in Figure \ref{fig:df_anno}, the Deepfake example reveals distinct differences in annotation approaches. Human annotations focus primarily on obvious visual cues like facial symmetry and cheek irregularities, but also incorrectly identify nose distortions. GPT-4o's description tends toward general stylistic observations about computer generation and animation-like qualities, lacking specific artifact identification. DD-VQA provides more structured observations about the eyes and mouth regions, correctly identifying texture patterns and blending artifacts, though still missing some key details.
Our FFTG's raw annotation demonstrates superior accuracy by precisely identifying the manipulated regions indicated by the forgery mask. It correctly pinpoints unusual texture patterns in the eyes and highlights blending artifacts around the eyes and mouth, while also detecting color distribution inconsistencies. This mask-guided approach helps avoid the hallucination of non-existent artifacts and ensures descriptions align with actual manipulation evidence.

\begin{figure*}[!t]
    \begin{center}
    
       \includegraphics[width=0.8\linewidth]{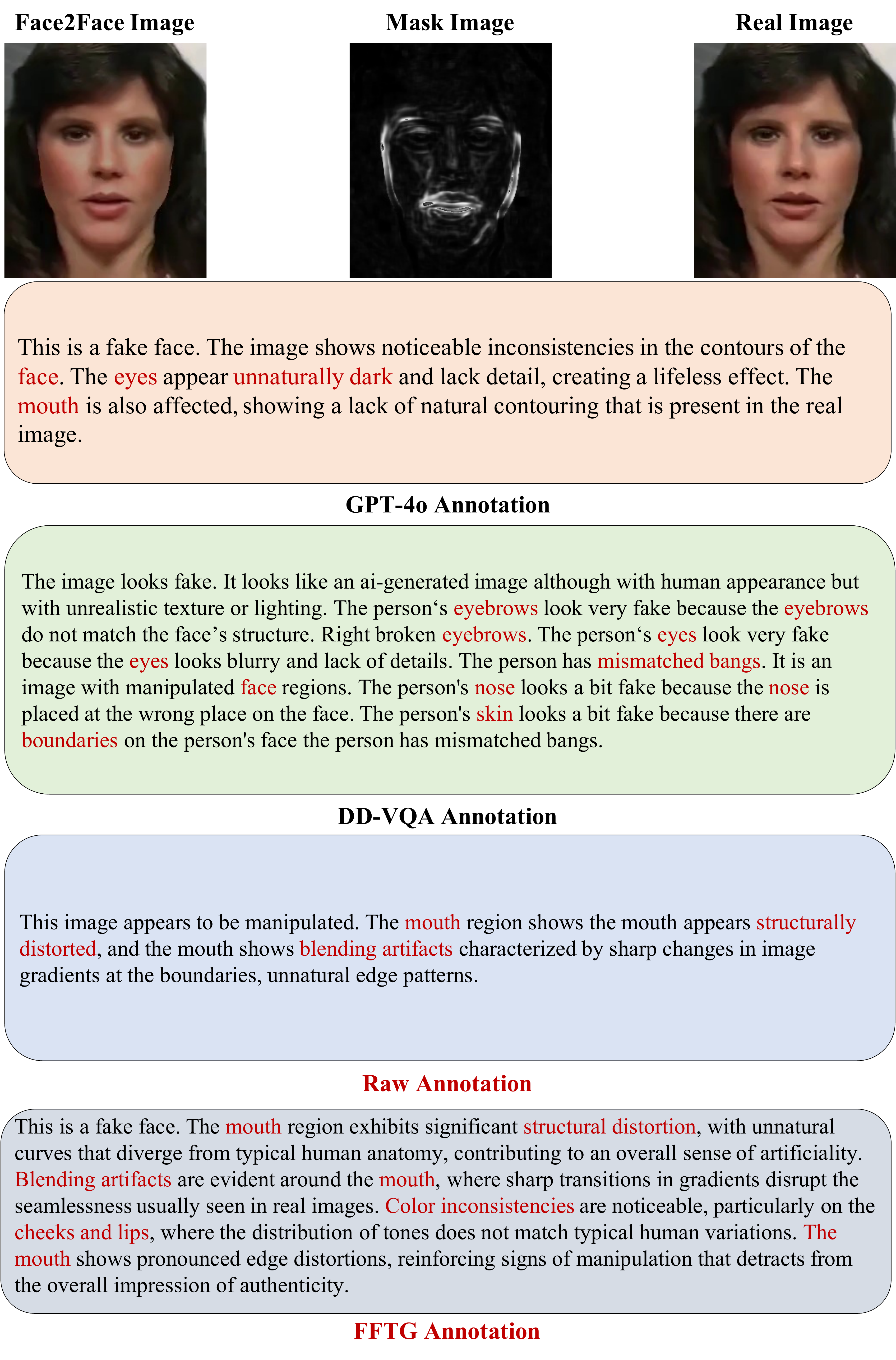}
    \end{center}
       \caption{Comparison of different annotation approaches on a Face2Face example.
       }
    \label{fig:f2f_anno}
\end{figure*}

For Face2Face manipulation (Figure \ref{fig:f2f_anno}), the human annotation correctly identifies the unnatural contouring and lighting around the face, particularly noting mouth region abnormalities. GPT-4o mentions various facial features including eyebrows and skin texture, but seems scattered in its focus. DD-VQA provides a more concise description focusing specifically on the structural distortion and blending artifacts in the mouth region. Our FFTG raw annotation shows the highest precision by accurately identifying structural distortions in the mouth area and highlighting specific artifacts like color inconsistencies and blending anomalies at region boundaries, which aligns well with the forgery mask's indication.

\begin{figure*}[!t]
    \begin{center}
    
       \includegraphics[width=0.8\linewidth]{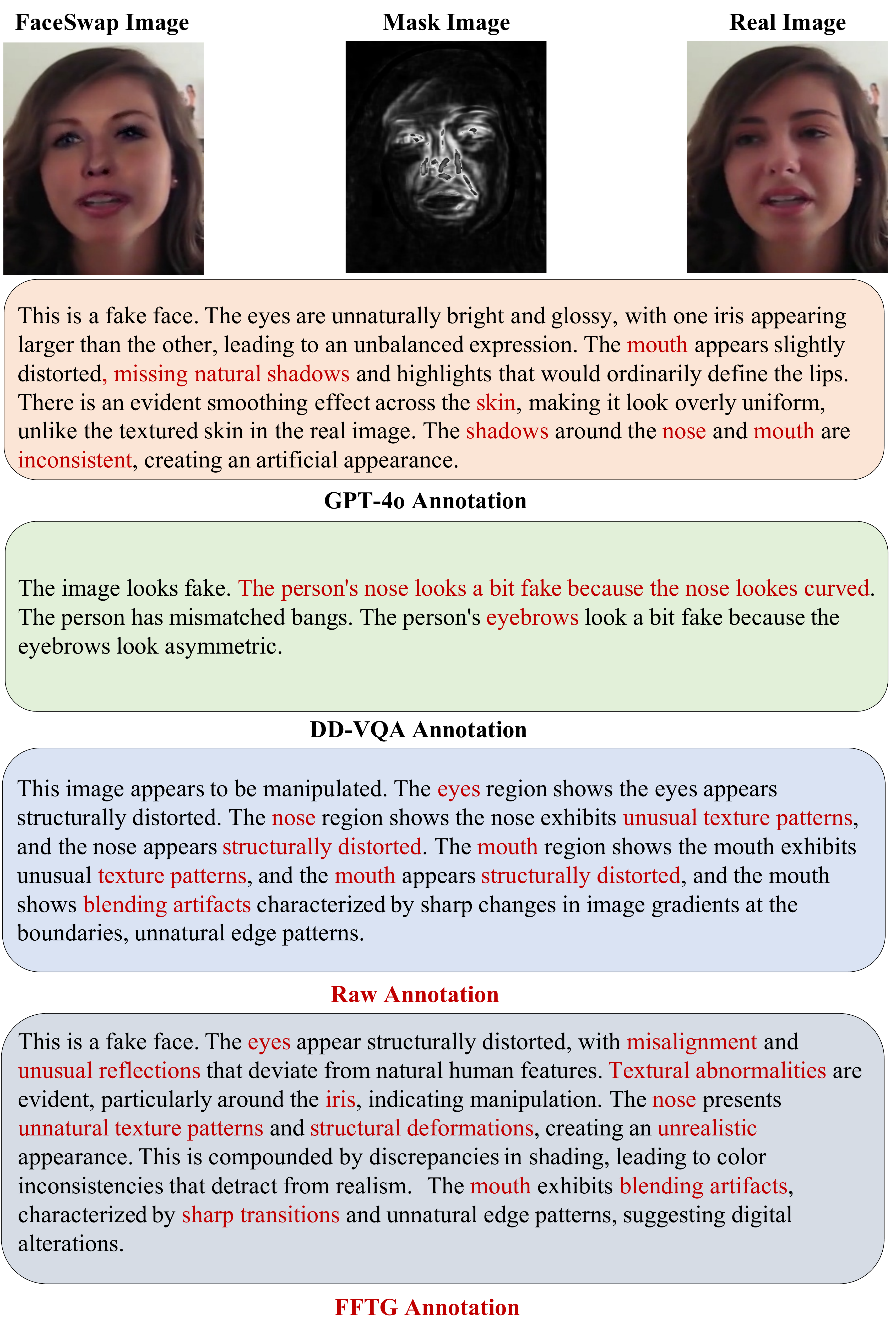}
    \end{center}
       \caption{Comparison of different annotation approaches on a FaceSwap example.
       }
    \label{fig:fs_anno}
\end{figure*}

In the FaceSwap example (Figure \ref{fig:fs_anno}), human annotation identifies unnatural brightness in the eyes and mouth distortions, along with skin smoothing effects. GPT-4o's description is notably limited, only mentioning curved nose and eyebrow asymmetry. DD-VQA provides more comprehensive detection, identifying structural distortions across eyes, nose, and mouth regions, with proper attention to blending artifacts. FFTG's raw annotation demonstrates superior precision by accurately capturing both the structural distortions and texture abnormalities in the eyes and nose regions, while also detailing the blending artifacts around the mouth, closely matching the forgery mask's indications.

\begin{figure*}[!t]
    \begin{center}
    
       \includegraphics[width=0.8\linewidth]{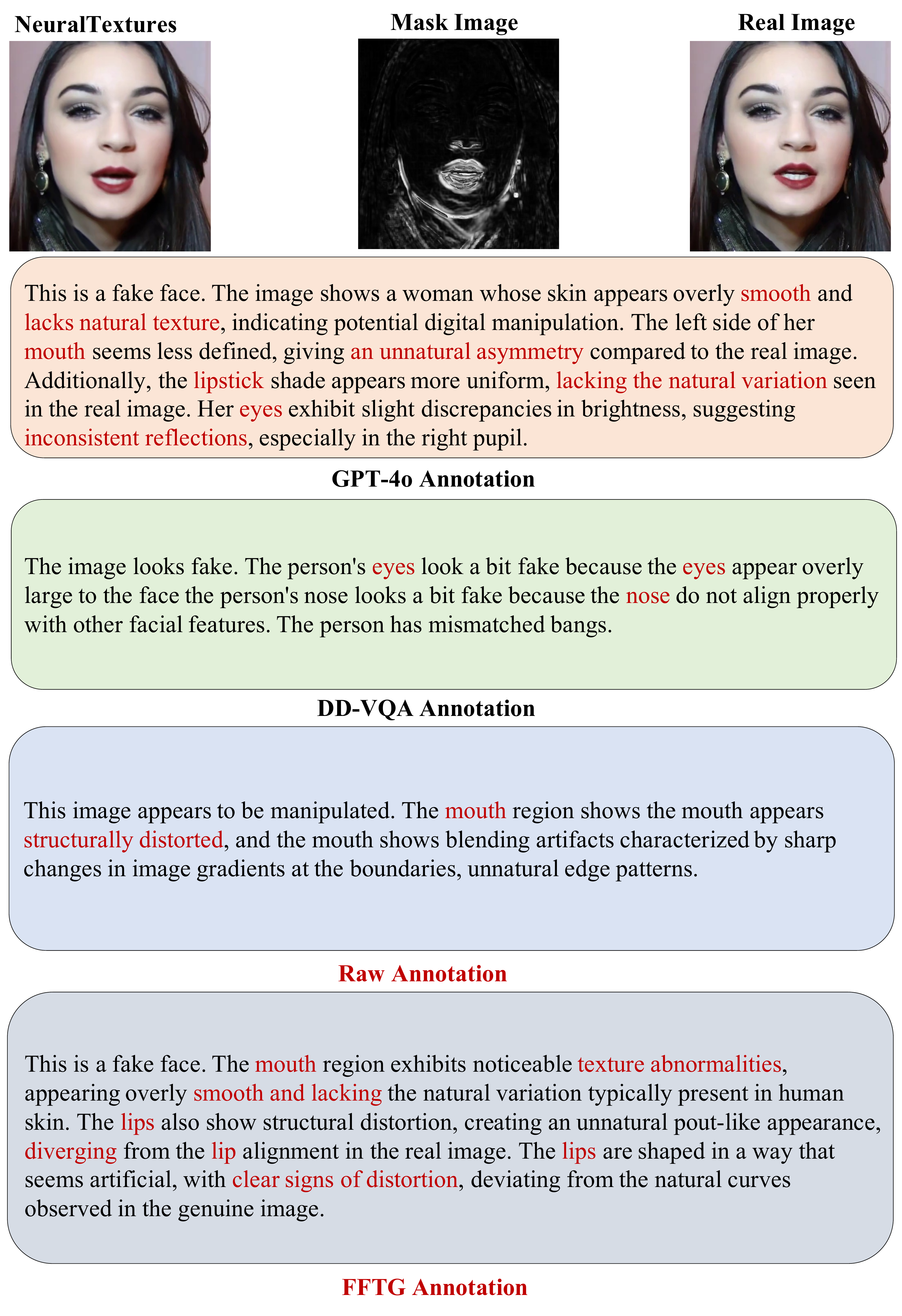}
    \end{center}
       \caption{Comparison of different annotation approaches on a NeuralTextures example.
       }
    \label{fig:nt_anno}
\end{figure*}

In the NeuralTextures example (Figure \ref{fig:nt_anno}), human annotation focuses on skin texture and asymmetry issues, particularly noting abnormalities in the mouth and lipstick regions. GPT-4o provides minimal observation, only mentioning eye and nose irregularities without specific details. DD-VQA maintains a focused description of the mouth region's structural distortions and blending artifacts. FFTG's raw annotation demonstrates the most precise detection by identifying specific texture abnormalities in the mouth region and structural distortions in the lip area, matching the forgery mask's indication of manipulation. The annotation particularly emphasizes unnatural texture patterns and deviations from natural curves, providing detailed evidence of manipulation.

Across all four manipulation types, FFTG consistently demonstrates superior accuracy in identifying and describing forgery artifacts, with its annotations closely aligning with the ground truth masks while providing detailed, artifact-specific descriptions that avoid hallucination.

\subsection{Visualizations of LLaVA Responses}


We demonstrate the effectiveness of FFTG annotations in improving multimodal language models' forgery detection capabilities through both quantitative evaluation and qualitative analysis. As shown in Table \ref{table:1}, our FFTG-enhanced LLaVA achieves superior performance across all metrics, with 95.84\% accuracy on FFpp and 75.00\% on the challenging Celeb-DF dataset, significantly outperforming models trained with DD-VQA annotations. More importantly, our model demonstrates higher precision (88.07\%) and recall (55.30\%) in identifying manipulation regions, indicating more accurate and reliable detection capabilities.

\begin{figure*}[!t]
    \begin{center}
    
       \includegraphics[width=0.9\linewidth]{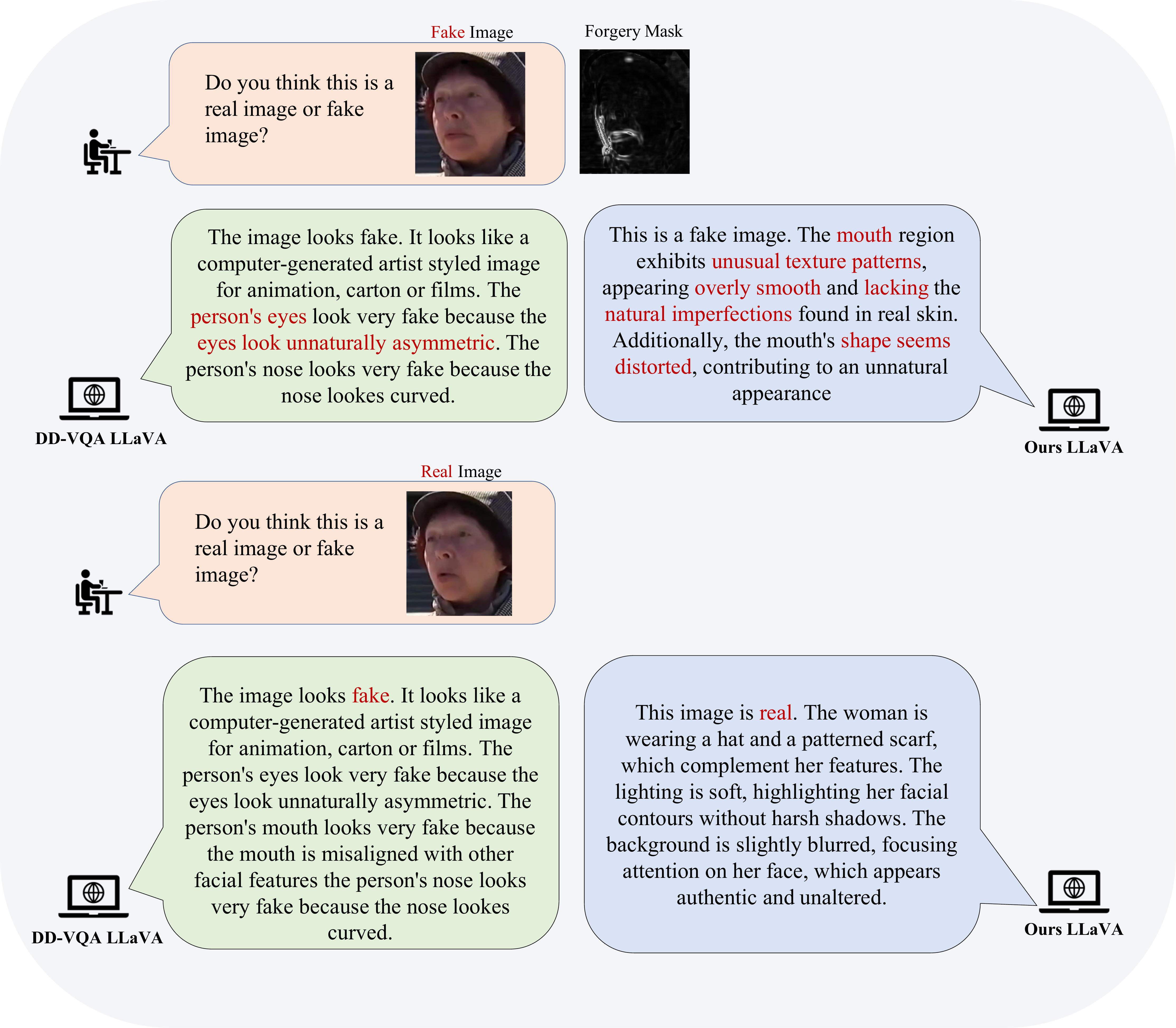}
    \end{center}
       \caption{Interactive dialogue comparison between DD-VQA-trained and FFTG-trained LLaVA models on both fake (top) and real (bottom) images.
       }
    \label{fig:llava}
\end{figure*}

This quantitative improvement is further illustrated through example dialogues in Figure \ref{fig:llava}. When presented with a challenging fake image, DD-VQA-trained LLaVA relies heavily on general stylistic observations about computer generation and animation-like qualities, focusing on superficial features like eye asymmetry and nose curvature. In contrast, our FFTG-trained LLaVA provides more precise and artifact-focused analysis, accurately identifying specific texture patterns in the mouth region and structural distortions that deviate from natural appearances. More importantly, when analyzing real images, while DD-VQA-trained LLaVA exhibits bias toward forgery detection with false positives, our model demonstrates better discrimination ability by correctly identifying authentic images and providing detailed natural features as supporting evidence. These qualitative examples, supported by the strong numerical results, demonstrate that FFTG's precise annotation guidance helps LLaVA develop more reliable and interpretable forgery detection capabilities.

\section{Prompt Details}
\subsection{Connectives of Raw Annotation}
\begin{figure*}[!t]
    \begin{center}
    
       \includegraphics[width=1.0\linewidth]{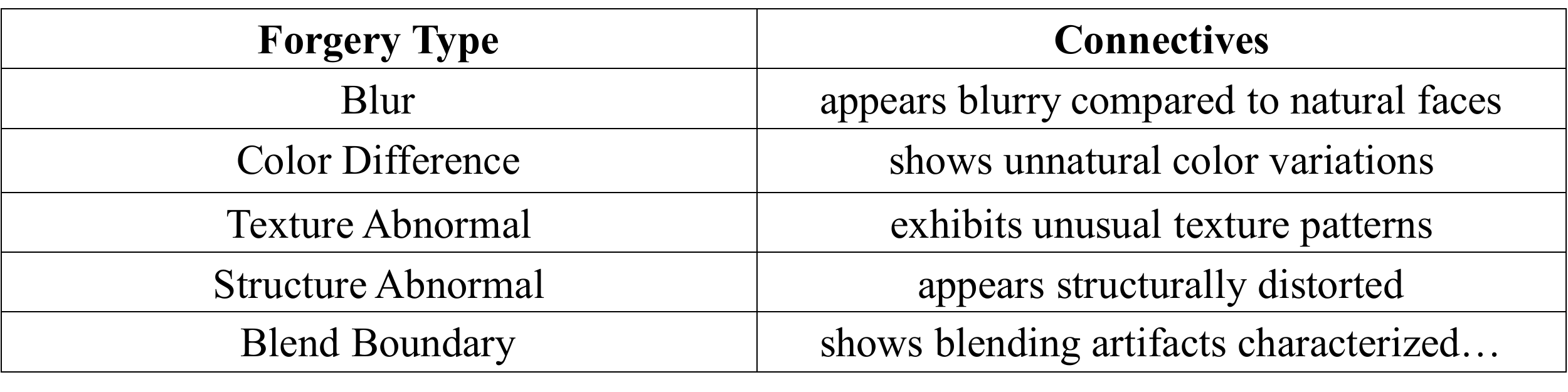}
    \end{center}
       \caption{Connective phrases used for different forgery types in raw annotation generation. Each phrase starts with a specific region token {region} (e.g., eyes, nose, mouth) followed by these connectives to form natural descriptions of detected artifacts.
       }
    \label{fig:raw_conn}
\end{figure*}

To enhance the naturalness and readability of raw annotations, we design specific connective phrases for each forgery type, as shown in Figure \ref{fig:raw_conn}. These connectives are used in conjunction with a region token (e.g., eyes, nose, mouth) to form complete, natural descriptions. For example, when blur is detected in the eye region, the annotation would read "the eyes appears blurry compared to natural faces". For blending artifacts, the base connective "shows blending artifacts characterized" is further enhanced with specific evidence phrases based on our detection metrics: "sharp changes in image gradients at the boundaries" when gradient discontinuity is detected, "unnatural edge patterns" for edge artifacts, and "unusual frequency patterns at the boundaries" for frequency domain abnormalities. These detailed characterizations help specify the exact nature of the blending artifacts detected.
This structured approach helps guide GPT in generating more accurate and contextually appropriate refined annotations while maintaining consistent terminology across different forgery types.

\subsection{Annotation Refinement Prompt}
\begin{figure*}[!t]
    \begin{center}
    
       \includegraphics[width=0.9\linewidth]{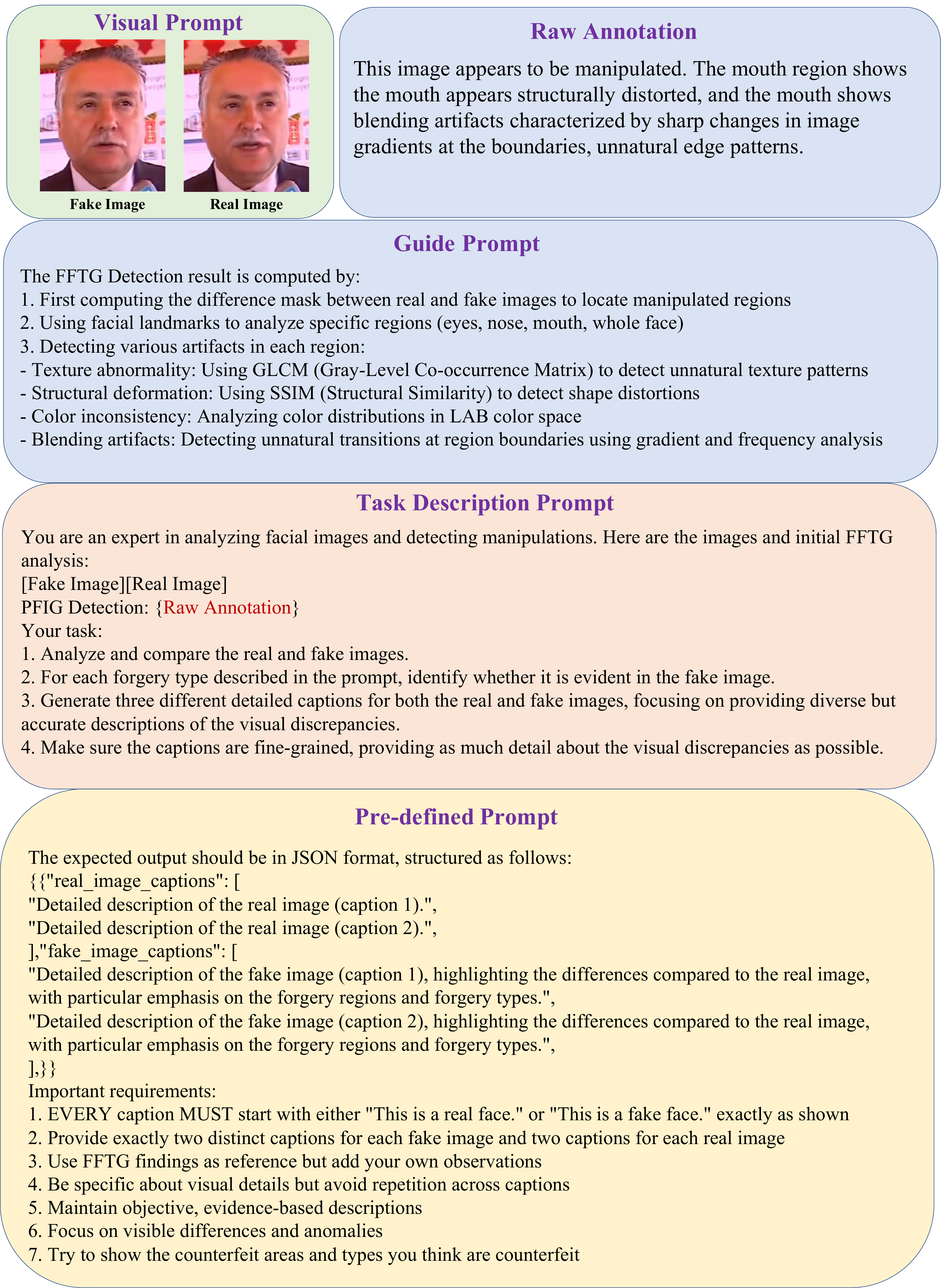}
    \end{center}
       \caption{Overview of FFTG prompting strategy for annotation refinement, consisting of Visual Prompt with paired images, Raw Annotation with dynamic descriptions, Guide Prompt explaining detection process, Task Description Prompt for analysis guidance, and Pre-defined Prompt for output format.
       }
    \label{fig:prompt}
\end{figure*}

To guide GPT in generating accurate and natural language annotations, we design four complementary prompts as shown in Figure \ref{fig:prompt}. The \textit{Visual Prompt} pairs fake and real images to enable direct visual comparison, helping GPT identify manipulation artifacts through contrast. For each case, we provide dynamically generated raw annotations that combine detected regions with corresponding connective phrases as initial guidance. The \textit{Guide Prompt} explains the FFTG detection process, including mask generation, region analysis, and specific criteria for detecting texture abnormalities, structural deformations, color inconsistencies, and blending artifacts, helping GPT understand the technical basis. The \textit{Task Description Prompt} establishes the expert analysis context and provides step-by-step instructions for comparing images and generating comprehensive descriptions. Finally, the \textit{Pre-defined Prompt} specifies the required JSON output format and key requirements to ensure consistent and focused annotations. 
This multi-faceted prompting strategy enables GPT to generate detailed, accurate descriptions while maintaining natural language expression and avoiding hallucination.

\end{document}